# Missile detection and destruction robot using detection algorithm

Md Kamrul Siam
Computer Science and Engineering
Military Institute of Science and Technology
Dhaka, Bangladesh
kamrulhossainsiam@gmail.com

Shafayet Ahmed
Computer Science and Engineering
Military Institute of Science and Technology
Dhaka, Bangladesh
ahmmedshafayet@gmail.com

Md Habibur Rahman
Computer Science and Engineering
Military Institute of Science and Technology
Dhaka, Bangladesh
Mahin.habib7192@gmail.com

Amir Hossain Mollah
Computer Science and Engineering
Military Institute of Science and Technology
Dhaka, Bangladesh
amir5153@gmail.com

**Abstract**- This research is based on the present missile detection technologies in the world and the analysis of these technologies to find a cost effective solution to implement the system in Bangladesh. The paper will give an idea of the missile detection technologies using the electro-optical sensor and the pulse doppler radar. The system is made to detect the target missile. Automatic detection and destruction with the help of ultrasonic sonar, a metal detector sensor, and a smoke detector sensor. The system is mainly based on an ultrasonic sonar sensor. It has a transducer, a transmitter, and a receiver. Transducer is connected with the connected with controller. When it detects an object by following the algorithm, it finds its distance and angle. It can also assure whether the system can destroy the object or not by using another algorithm's simulation.

# TABLE OF CONTENT

CHAPTER 1

# INTRODUCTION

## 1.1 Overview

Bangladesh, a South Asian peaceful country, has been developing its economy gradually though it has lots of obstacles in daily politics, good governance and democracy in the country.

However, Bangladesh has been improving its defense capability within its small economy to protect sovereignty and independence and ensure international dignity by taking part peace keeping mission in foreign countries under United Nation. Historically Bangladesh has been procuring weapons from China since 1975 when a diplomatic relation had been made with China. Bangladesh purchases 80% weapons from China because of its decent quality, lower price and easy to use. In addition, sometime Bangladesh procures weapons from other countries like Russia, Serbia, Italy, and so on. But Bangladesh does not purchase weapons from USA, UK, France, Germany because of its high prices which is not capable of Bangladesh defense budget as Bangladesh is a small economic country.

Bangladesh spends 6.3% on defense budget of its total budget, so its maintenance and other expenditures end up the major part of the defense budget. Bangladesh has no large defense budget for procuring more powerful weapons from other countries, and has no capability to produce such weapons lack of technology and experts.

Since our defense budget is not that much bigger, we have to concentrate on producing our own system in a low cost budget. Since our neighboring countries have Ballistic missile and Tactical missiles including Nuclear weapons, to maintain the power balance and countries safeguard, we have to achieve Missile Defense System technology of our own in a low cost budget.

## 1.2  Missile Defense System

Missile defense is a system, weapon, or technology involved in the detection, tracking, interception, and destruction of attacking missiles. Originally conceived as a defense against nuclear-armed intercontinental ballistic missiles (ICBMs), its application has broadened to include shorter-ranged non-nuclear tactical and theater missiles. [1]

Missile defense systems are a type of missile defense intended to shield a country against incoming missiles, such as intercontinental ballistic missile (ICBMs) or other ballistic missiles. The United States, Russia, India, France, Israel and China have all developed missile defense systems.

The term "Missile defense system" broadly means a system that provides any defense against any missile type (conventional or nuclear) by any country.

Any mechanism which can detect and then destroy a missile before it can cause any harm is called a missile defense system (MDS).

The role of defense against nuclear missiles has been a heated military and political topic for several decades. However, missile defense is no longer limited to interception of strategic nuclear weapons. The gradual development and proliferation of missile technology has blurred the line between the technologies for the interception of tactical missiles (usually short to intermediate range with non-nuclear payloads) and the interception of strategic missiles (usually long ranged with nuclear payloads). High-performance tactical ballistic missiles carrying non-nuclear payloads now have the ability to affect strategic balance in conflict zones. Likewise, high-performance tactical missile defense systems can now influence force deployment strategies.

## 1.3  Contribution

In this work, A SONAR based RADAR system is displayed by incorporating Smoke sensor and Metal detection sensor with it for the Detection of missile. For tracking of missile, we have used pure pursuit Algorithm. The purpose of this paper is to help our country to build a missile defense system within a low budget which can be accommodated in our economy. The whole paper's focus is divided into following stages:

    a. Missile Launch Detection System.

    b. Missile Tracking System

    c. Signal Processing System.

    d. Target engagement and Destruction.

## 1.4    Organization of the Paper

Chapter 2: Literature Review

Chapter 3: Missile Detection System

Chapter 4: Missile Tracking System

Chapter 5: Target engagement and Destruction

Chapter 6: Concludes the paper

# CHAPTER 2

# LITERATURE REVIEW

## 2.1　History of Missile Defense System

In the 1950s and 1960s, the term meant organization against strategic (usually nuclear-armed) missiles. The engineering mostly concentrated around finding offensive launch events and following inward ballistic missiles, but with controlled power to actually defend against the missile. The SOVIET UNION achieved the oldest nonnuclear intercept of a ballistic missile warhead by a missile at the Sary Shagan antiballistic missile defense experiment formation on 4 March 1961.

Throughout the 1950s and 1960s, the United States Project Nike air defense program focused initially on bombers, then ballistic missiles. In the 1950s, the first United States anti-ballistic missile system was the Nike Hercules, which had a limited ability to intercept incoming ballistic missiles, although not ICBMs. This was followed by Nike Zeus, which using a nuclear warhead could intercept ICBMs. However it was feared the missile's electronics may be vulnerable to x-rays from a nuclear detonation in space. A program was started to devise methods of hardening weapons from radiation damage.[2] By the early 1960s the Nike Zeus was the first anti-ballistic missile to achieve hit-to-kill (physically colliding with the incoming warhead).[1]

In the early 1980s, technology had matured to consider space based missile defense options. Precision hit-to-kill systems more reliable than the early Nike Zeus were thought possible. With these improvements, the Reagan Administration promoted the Strategic Defense Initiative, an ambitious plan to provide a comprehensive defense against an all-out ICBM attack. Reagan established the Strategic Defense Initiative Organization (SDIO), which was later changed to the Ballistic Missile Defense Organization (BMDO). In 2002, BMDO's name was changed to its current title, the Missile Defense Agency (MDA). See National Missile Defense for additional details. In the early 1990s, missile defense expanded to include tactical missile defense, as seen in the first Gulf War. Although not designed from the outset to intercept tactical missiles, upgrades gave the Patriot system a limited missile defense capability. The effectiveness of the Patriot system in disabling or destroying incoming Scuds was the subject of Congressional hearings and reports in 1992 [3].

In the last of 1990s, and old 2000s, the bare of defense against cruise missiles became many more important with the new Bush Administration. In 2002, President George W. Bush withdrew the US from the Anti-Ballistic Missile Treaty, allowing more development and testing of ABMs under the agency of missile defense, and allowing for deployment of interceptor vehicles beyond the only place allowed under the pact.

There are still technological hurdles to an effective defense against ballistic missile attack. The United States National Ballistic Missile Defense System has come under scrutiny about its technological feasibility. Intercepting midcourse (rather than launch or reentry stage) ballistic missiles traveling at several miles per second with a "kinetic kill vehicle" has been characterized as trying to hit a bullet with a bullet. Despite this difficulty, there have been several successful test intercepts and the system was made operational in 2006, while tests and system upgrades continue. Moreover, the warheads or payloads of ballistic missiles can be concealed by a number of different types of decoys. Sensors that track and target warheads aboard the kinetic kill vehicle may have trouble distinguishing the "real" warhead from the decoys, but several tests that have included decoys were successful. Nira Schwartz's and Theodore Postol's criticisms about the technical feasibility of these sensors have led to a continuing investigation of research misconduct and fraud at the Massachusetts Institute of Technology[4].

## 2.2 Field of Work

Our work is directed towards a missile detection and defense System. The System of the present work comprises a missile, a missile launch detection System, a missile tracking System, and a signal processing System capable of receiving Said tracking Signal, calculating an intercept trajectory for a missile to intercept a missile, and further capable of outputting an intercept trajectory program to a missile.

## 2.2 Description of the Prior Art

The trajectory of a long range missile will follow an arc like path. The initial one third of the arc comprises the path of the missile from immediately after it is fired as it ascends toward its target along the arc like trajectory. The middle third portion of the arc comprises the Zenith of the missile's trajectory, when the missile trajectory Switches from ascending to descending. The final third of the arc comprises the missile's descent toward, and impact with, its target. Missile detection and defense Systems may be divided into categories based upon the intended portion of the missile's arc trajectory where interception is intended to occur. This method of classification is referred to herein as "trajectory trisection."

Prior art Missile Defense systems have been directed toward intercepting missiles as they are descending toward their target in the final phase of the trajectory trisection category. One of the benefits of such a system is that Significant time is available to track the incoming ballistic missiles, calculate their trajectory, and distinguish decoys from actual missiles.

One of the major drawbacks of Such a system is that the incoming missile is relatively close to its target by the time Such a Missile Defense System launches an interceptor missile. If the interceptor missile misses or experiences a malfunction, inadequate time is left to take alternate defensive measures. In Such a Scenario, if the incoming missile contains a thermonuclear warhead, large Scale destruction and radioactive contamination will result. If the incoming missile detonates near a population center, millions of lives may be lost and billions of dollars in property damage is likely to result. Thus, the risks associated with Such a System appear to far exceed the benefits. Intercepting a missile at a point relatively close to the target presents danger to people and property in the target vicinity from falling debris resulting from a Successful missile interception.

Other Missile Defense systems are directed toward intercepting missiles in the middle phase of their trajectory trisection category. One Such System is disclosed in Jun. 20, 2000 documents published by the United States Department of Defense ("USDoD'). These documents disclose a Missile Defense System intended to launch a land based "kill vehicle" intended to intercept an incoming missile in midcourse. Under the presidential administration of William Clinton, this system represented the choice of the USDoD for the National Missile Defense ("NMD) system. Such a System provides less time to evaluate the trajectory of the incoming missile than a final phase trajectory trisection System. Such a System pro vides more time to evaluate the trajectory of the incoming ballistic missile than an initial phase trajectory trisection System.

Missile Defense Systems may be particularly well Suited for defending against Small Scale missile attacks Such as those which the USDoD believes will be possible by the year 2005 from Small nations, Such as North Korea. Such nations are referred to by the USDoD as "rogue nations" in a Jan. 20, 1999 DoD News Briefing by Secretary of Defense William S. Cohen, published by the USDoD. ABM systems may also be classified according to whether the interceptor missile, detection Systems or control Systems is land based or nonland based. Land or ground based systems are disclosed in the Millard patent, in USDoD news briefings describing the NMD system, and in U.S. Pat. No. 5,340,056 to Guelman et al. and U.S. Pat. No. 5,464,174 to Laures.

There are Several disadvantages to land based Systems. One disadvantage of land based Systems is the limited geographic area which they can cover. A simple land based System intended to protect the population centers and military installations of the west coast of the lower 48 states of the United States would have to cover a coastline Stretching approximately 1100 miles, from the Mexican border to the Canadian border. If Such a System is to be a final phase trajectory trisection System, Such as, Multiple land based missile sites must be employed to protect the intended target zone. Also, Machine learning approaches can be used to design system and predict the behavior of missile[18].

# CHAPTER 3

# MISSILE DETECTION SYSTEM

## 3.1 Introduction

'To see and yet not be seen', that is the character of modern warfare. Significantly, current and future weapon systems are capable of easily wiping out targets, in the air or on the ground, once detected and identified. This of course places incentives upon all parties, incentives to locate and reliably recognize the enemy if possible without being detected in the process.

Missile Detection System (MDS) is the first part of the Anti-Missile System. The Missile Detection System is the combination of both Hardware and Software. The Hardware consists of multiple sensors which detect different parameters of an incoming Missile. The detected signals by the sensors are eventually passed to the detection software for further processing. Finally, the detection software analyzes the data given by the sensors to find out actual information about the incoming Missile. Generally, sensors used in the Missile Detection System are of two categories: Ground Based Sensors and Space Based Sensors. An approaching Missile is a heated metal body which carries warheads. It also bears some smoke with it. Thus a system composed of metal detector, smoke sensor and heat sensor can identify the existence of an incoming Missile. The Missile Detection System is commonly known as RADAR. Different types of RADAR technologies are available in the present world.

## 3.2 Steps of Missile Detection System

The Missile Detection System includes the following steps:

1. Detection of a moving object in air and its speed.

2. Detection of the presence of smoke and fire in the object.
3. Distinguishing the Missile from other moving objects in air.
4. If it is a Missile then, Calculating the angle and the velocity of the flight.

### 3.2.1 Detection of a moving object in air

Detection of a moving object in air is done by using different type of sensors. Most commonly used ones are: SONAR sensor, Infrared sensor and Electro-Optical Sensor. These sensors can detect the presence of any moving object as well as the speed of the object. But the detection of a moving object does not necessarily decide that this is a Missile. That decision requires some more functions of the RADAR system.

### 3.2.2 Detection of the presence of smoke and fire in the object

After the detection of the moving object, the next step is to detect the presence of any smoke and fire in the object. This detection cannot be done by the normal fire and smoke sensors. The sensors need to cover a high range. So, this system uses the high range smoke and fire detectors which uses infrared.

### 3.2.3 Distinguishing the Missile from other moving objects in air

The next step of detection is to distinguish between a Missile and the other moving objects in air. This is the most important and sophisticated part of the detection system. Other moving objects in the air may also generate smoke and heat. So, the system must be able to distinguish between the objects like fighter aircraft, bird and a Missile. The System performs this function by allocating different frequency bands for different objects.

### 3.2.4 Calculating the angle and the velocity of the flight

After getting the confirmation that the object is a Missile, the next objective of the detection system is to calculate the velocity and the angle of the flight. This is done by some calculations using the data provided by the sensors. The detail procedure will be discussed in the next chapters.

## 3.3 Missile Defense RADAR Systems and Sensors

Strategic and Tactical missile detection mainly consists of two functions. Firstly, The detection of the launching of the missile and the burning of the propellant during the acceleration in the trajectory. Secondly, the tracking of the missile to find out where the missile is targeted. When multiple launch is done then the number of the missiles are counted and tracked. There are number of RADAR systems and advanced sensor technologies available in the present world. Though Radar is a potent device, especially in its advanced forms, it has a fundamental impotence as it requires that the communicator illuminate the target with energy. In doing so it identifies itself and betrays its location, not to utter of its vulnerability to deceptive. Electro-optical (EO) sensors, on the opposite side, do not have these deficiencies, as they are all passive, sensing energy emitted by or mirrored off the aim itself. We will discuss two of them which are the most important and effective ones:

1. Pulse-Doppler Radar
2. Missile Defense Infrared and Electro-optical sensors

## 3.4 Pulse-Doppler Radar

A 'Pulse-Doppler Radar' is a radar system that identifies the range of a target using pulse-timing technology and use the Doppler effect of the returned signal to determine the missiles velocity. It combines the facilities of pulse radar and continuous-wave radar, which were previously separate due to the complexity of the electronics. [5]

Pulse-Doppler system was first used on fighter aircraft starting in the 1960s. Initial radars used pulse-timing in order to accomplish range and the position of the antenna to decide the bearing. This functioned when the radar antenna was not pointed downwards; in that framing the reflection off the ground overcome any returns from opposite substances. As the ground moves at the same speed but in the reverse direction of the aircraft, Doppler techniques allows the return from the ground to be filtered. This gives pulse-Doppler radars "look-down/shoot-down" potentiality. An alternative asset in military radar is to lessen the transmitted power while achieving agreeable performance for more safety of silent radar.

### 3.4.1 Pulse Radar Operation

Most radar systems are pulsed, meaning that the radar will transmit a pulse and then listen for receive signals, or echoes. This avoids the problem of a sensitive receiver trying to operate simultaneously with a high power transmitter.

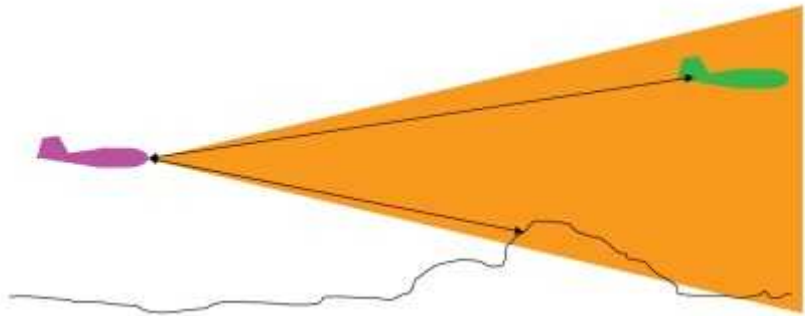

Figure 3.1: A representation of basic radar operation.

The pulse width or duration is an important factor. The radars operate by "binning" the receive signals. The receive signal returns are sorted into a set of bins by time of arrival relative to the transmit pulse. The time interval is in proportion to the round-trip distance to the object(s) reflecting the radar waves. By checking the receive signal strength in the bins, the radar can sort the returns across the different bins, which correspond to different ranges. This can be performed while scanning across desired azimuths and elevations.

Having many range bins allows more precise range determinations. A short duration pulse is likely to be detected and mapped into only one or two range bins, rather than being spread over many bins. However, a longer pulse duration or width allows for a greater amount of signal energy to be transmitted and a longer time for the receiver to integrate the energy. This means longer detection range. In order to optimize for both fine range resolution and long range detection, radars use a technique called pulse compression.

### 3.4.2 Pulse Compression

The goal of pulse compression is to transmit a long duration pulse of high energy, but to detect a short duration pulse in order to localize the receive filter output response to one or at most two radar range bins. Early radars accomplished this by transmitting a signal with linear frequency modulation. The pulse would start at a low frequency sinusoid and increase the frequency over the duration of the radar pulse. This is referred to a "chirp". A special analog filter is used at the receive end, with non-linear phase response. This filter has a time lag that decreases with frequency. When this rate of time lag decrease is matched to the rate of increase in the chirp, the result is a very short, high amplitude output from the filter. The response of the pulse detection has been "compressed".

Digital radars also perform pulse compression, but using matched filter. The transmitted radar pulse uses a pseudo random sequence of phase modulations and is detected using a filter matched to that same sequence, then the resulting output will match only when the stored sequence matches the received sequence. The timing resolution of the receive signal is equal to the transition time of the phase changes, which can be very rapid. This detection method can also filter out undesired signals that do not match the stored sequence.

### 3.4.3 Pulse Repetition Frequency

The pulse repetition frequency (PRF) is the rate of transmitting pulses, on which receive processing is then performed. The higher the PRF, the greater the average power the radar is transmitting (assuming the peak power of each pulse is limited by the transmit circuitry) and the greater the detection range. A high or fast PRF also allows for more rapid detection and tracking of objects, as range measurements at a given azimuth and elevation can be performed during each PRF interval. A high PRF is also a disadvantage, which means it has a limited distance to perform unambiguous determination of range.

Range to target is measured by round trip delay in the received echo. It is the speed of light multiplied by the time delay and divided by two to account for the round trip.

$$R_{measured} = v_{light}\, t_{delay} / 2$$

The maximum range that can be unambiguously detected is limited by the PRF. This is more easily seen by example. If the PRF is 10 kHz, then there is 100 us between pulses. Therefore, all return echoes should ideally be received before the next transmit pulse. This range is simply found by multiplying the echo delay time by the speed of light and dividing by two to account for the roundtrip.

$$R_{maximum} = (3 \times 10^8 \text{ m/s}) (100 \times 10^{-6} \text{ sec}) / 2 = 15 \text{ km}$$

Suppose the radar system sorts the returns into 100 range bins, based upon the time delay of reception. The range resolution of this radar system is then 0.15 km, or 150 meters. However, there may be returns from distances beyond 15 km. Suppose that a target aircraft #1 is five km in the distance and target aircraft #2 is twenty-one km in the distance.

Target aircraft #1 will have a delay of:

$$t_{delay} = 2\, R_{measured} / v_{light} = 2\, (5 \times 10^3) / 3 \times 10^8 = 33 \text{ us}$$

Target aircraft #2 will have a delay of:

$$t_{delay} = 2\, R_{measured} / v_{light} = 2\, (21 \times 10^3) / 3 \times 10^8 = 140 \text{ us}$$

The first target return will be mapped into the 33rd out of 100 range bin and the second target to 40th range bin. This is called a range ambiguity. The target(s) which are within the 15 km range are said to be in the unambiguous range. This is analogous to the sampling rate. The range ambiguity is analogous to aliasing during the sampling process (see Figure 3).

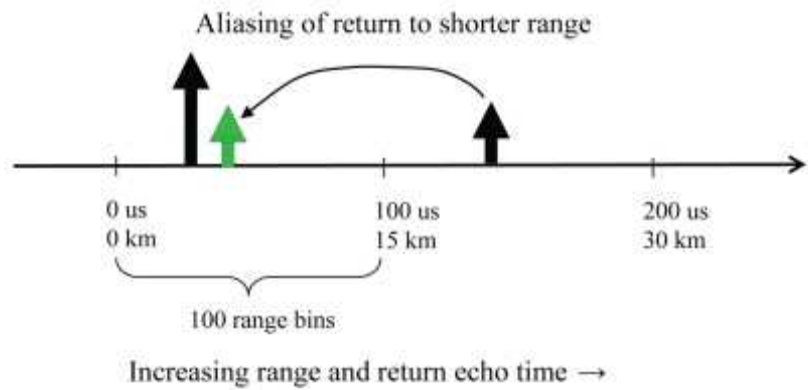

Figure 3.2: A pictorial representation of aliasing that causes range ambiguity.

The obvious solution is to use a lower PRF. This approach works, but the benefits of the high PRF are lost. Another solution to this problem is to transmit different pulses at each PRF interval. However, this has the downside of complicating the receiver, as it must now use multiple matched filters at each range bin and at each azimuth and elevation. This will effectively double the rate of digital signal processing required for each separate transmit pulse and matched filter pair used.

Typically, most radar systems will vary the PRF, depending on the operational mode of the radar.

### 3.4.4 History of Pulse-Doppler Radar

The initial radar systems failed to operate as expected. The reason was traced to Doppler effects that reduce performance of systems not planned to account for moving things. Fast-moving things cause a phase-shift on the transmit pulse that can harvest signal cancellation. Doppler has maximum negative effect on moving target pointer systems, which must use opposite phase shift for Doppler reparation in the detector. Doppler weather effects were also found to degrade conventional radar and moving target indicator radar, which can mask aircraft reflections. This phenomenon was adapted for use with weather radar in the 1950s after declassification of some World War II systems.

Pulse-Doppler radar was developed during World War II to overwhelmed limitations by increasing pulse recurrence frequency. This required the development of the klystron, the traveling wave tube, and solid state devices. Pulse-Doppler is incompatible with other high power microwave amplification devices that are not coherent.

Early examples of military systems include the AN/SPG-51B developed during the 1950s specifically for the purpose of operating in hurricane conditions with no performance degradation. [5]

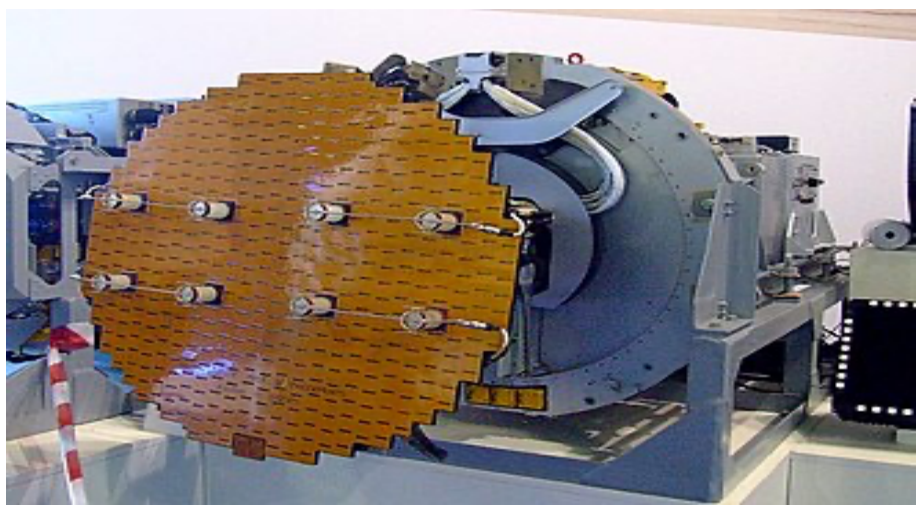
Figure 3.3: Airborne pulse-doppler radar antenna

### 3.4.5 Principle

Pulse-Doppler quantify the distance of the object by measuring the elapsed period between sending a pulse of radio energy and receiving a reflection of the object. Radio wave roaming at the velocity of light, so the distance to the object is the elapsed time multiplied by the speed of light, divided by two - there and back.

Pulse-Doppler radar is based on the Doppler effect , where movement in range produces frequency shift on the signal reflected from the target.

The relationship between wavelength and frequency is:

$$\lambda = v / f$$

where:
f = wave frequency (Hz or cycles per second)
λ = wavelength (meters)
v = speed of light (approximately $3 \times 10^8$ meters/second)

What happens in a radar system is that the pulse frequency is modified by the process of being reflected by a moving object. Consider the transmission of a sinusoidal wave. The distance from the crest of each wave to the next is the wavelength, which is inversely proportional to the frequency.

Each successive wave is reflected from the target object of interest. When this object is moving towards the radar system, the next wave crest reflected has a shorter round trip distance to travel, from the radar to the target and back to the radar. This is because the target has moved closer in the interval of time between the previous and current wave crest.

As long as this motion continues, the distance between the arriving wave crests is shorter than the distance between the transmitted wave crests. Since frequency is inversely proportional to wavelength, the frequency of the sinusoidal wave appears to have increased. If the target object is moving away from the radar system, then the opposite happens. Each successive wave crest has a longer round trip distance to travel, so the time between arrival of receive wave crests is lengthened, resulting in a longer (larger) wavelength, and a lower frequency.

This effect only applies to the motion relative to the radar and the target object. If the object is moving at right angles to the radar there will be no Doppler frequency shift. An example of this would be airborne radar directed at the ground immediately below the aircraft. Assuming level terrain and the aircraft is at a constant altitude, the Doppler shift would be zero, as there is no change in the distance between the plane and ground.

If the radar is ground-based, then all Doppler frequency shifts will be due to the target object motion. If the radar is a vehicle or airborne-based, then the Doppler frequency shifts will be due to the relative motion between the radar and target object.

This can be of great advantage in a radar system. By binning the receive echoes both over range and Doppler frequency offset, target speed as well as range can be determined. Also, this allows easy discrimination between moving objects, such as an aircraft or vehicle, and the back ground clutter, which is generally stationary.

$$\text{Doppler frequency} = \frac{2 \times \text{transmit frequency} \times \text{range velocity}}{C}.$$

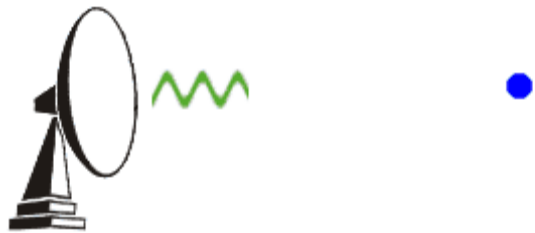

Figure 3.4: Principle of pulsed radar

### 3.4.6 Pulsed Frequency Spectrum

For this to be of any use, the Doppler shift must be measured. First, the spectral representation of the pulse must be considered.

The frequency response of an infinite train of pulses is composed of discrete spectral lines in the envelope of the pulse frequency spectrum. The spectrum repeats at intervals of the PRF.

What is important is that this will impose restrictions on the detectable Doppler frequency shifts. In order to unambiguously identify the Doppler frequency shift, it must be less than the PRF frequency. Doppler frequency shifts greater than this will alias to a lower Doppler frequency. This ambiguity is similar to radar range returns beyond the range of the PRF interval time, as they alias into lower range bins.

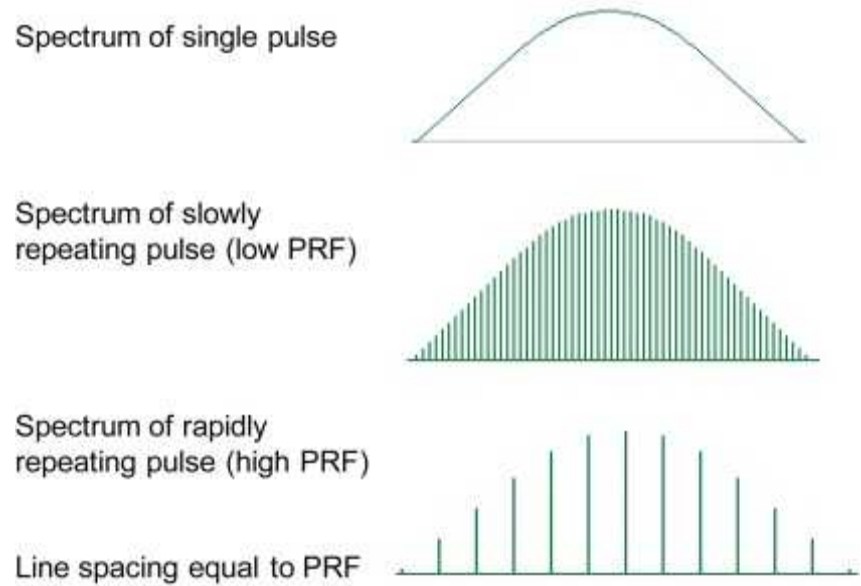

Figure 3.5. Pulse frequency spectrum

Doppler frequency detection is performed by using a bank of narrow digital filters, with overlapping frequency bandwidth (so there are no nulls or frequencies that could go undetected). This is done separately for each range bin. Therefore, at each allowable range, Doppler filtering is applied. Just as the radar looks for peaks from the matched filter detector at every range bin, within every range it will test across the Doppler frequency band to determine the Doppler frequency offset in the receive pulse.

### 3.4.7  Benefits of the Pulse-Doppler Radar

1. Rejection speed can be nominated on pulse-Doppler missile-detection systems. Therefore, nothing under the speed will be detected. A one degree antenna beam light up millions of square feet of terrain at 10 miles (16 km) range, and this produces thousands of detections at or below the horizon if Doppler is not used. [5]

2. In airborne pulse-Doppler radar, the velocity threshold is offset by the speed of the aircraft relative to the ground.

3. Surface reflections seem in almost all radar. Ground clutter generally appears in a circular region within a radius of about 25 miles (40 km) close ground-based radar. This distance extends much more in airborne and space radar. Clutter outcomes from radio energy being reflected from the earth surface, buildings, and vegetation. Clutter includes weather in radar envisioned to detect and report aircraft and spacecraft. [6] Clutter creates a vulnerability region in pulse-amplitude time-domain radar. Non-Doppler radar systems cannot be pointed directly at the ground due to excessive false alarms, which overwhelm computers and operators. Sensitivity must be reduced near clutter to avoid overload. This concern begins in the low-elevation region several beam widths above the horizon, and extends downward. This also exists throughout the volume of moving air associated with weather phenomenon.

4. Pulse-Doppler radar corrects this as follows:

   a. Allows the radar antenna to be pointed directly at the ground without overwhelming the computer and without reducing sensitivity.
   b. Fills in the susceptibility region associated with pulse-amplitude time-domain radar for small object detection near terrain and weather.
   c. Increases detection range by 300% or more in comparison to moving target indication (MTI) by improving sub-clutter visibility. [7]

### 3.4.8 Demerits of the Pulse-Doppler Radar

1. Uncertainty processing is required when target range is above the red line in the graphic, which rises scan time. Scan time is a acute factor for some systems because vehicles moving at or above the speed of sound can travel one mile (1.6 km) every few seconds, like the Exocet , Harpoon, Kitchen, and Air-to-air missile. The maximum time to scan the entire volume of the sky must be on the order of a dozen seconds or less for systems operating in that environment.

2. Pulse-Doppler radar by itself can be too slow to cover the entire volume of space above the horizon unless fan beam is used. This approach is used with the AN/SPS 49(V)5 Very Long Range Air Surveillance Radar, which costs elevation measurement to gain speed.[8]

3. Pulse-Doppler antenna motion must be slow enough so that all the return signals from at least 3 different PRFs can be processed out to the maximum anticipated detection range. This is known as dwell time. [9] Antenna motion for pulse-Doppler must be as slow as radar using MTI.

4. Search radar that embrace pulse-Doppler are usually twin mode because best overall performance is achieved when pulse-Doppler is used for areas with high false alarm rates (horizon or below and weather), while conventional radar will scan faster in free-space where false alarm rate is low (above horizon with clear skies).

5. The antenna type is a vital attention for multi-mode radar because unwanted phase shift introduced by the radar antenna can degrade performance measurements for sub-clutter visibility.

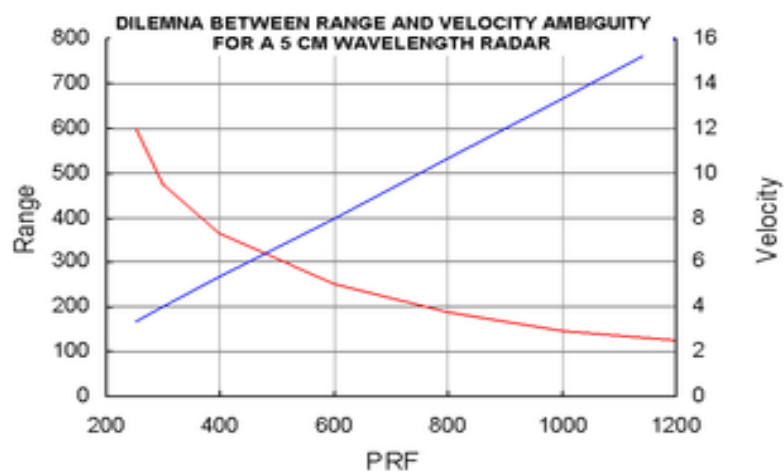

Figure 3.6: Maximum range from reflectivity (red) and unambiguous Doppler velocity range (blue) with a fix pulse repetition rate

## 3.5     Missile Defense Infrared and Electro-optical (EO) Sensors

The missile detection is done by the remote sensing of visual, ultraviolet and infrared sources. The remote sensing of these sources currently engage the electro-optical detection system. The Systems are riding on various stages. These stages are maintained by ships, aircrafts and satellites. The electro-optical or optical sensor is mainly a 'Telescope' which is used for collecting photons which signify the target and a 'Detector' which is used for converting the collected photons into electrons and thus an electronic current. Thus the electronic current provides the information about the target.

Generally, strategic and tactical missile launch, detection and tracking systems use the infrared surveillance method which is embedded in the satellite sensors. The satellite sensors are much more effective as it covers much larger surveillance area. A large surveillance area is required to correctly identify the location of the launching and also the location of the place where the missile is likely to hit.

This predisposition toward EO target identification and EO guided weapons does appear to be accelerating. The newest phrase is the Focal Plane Array (FPA), a small slab of Semiconductor that behaves as a single chip TV imagery figure, infrared or visual. FPAs are more compact, robust and certain than conventional vision supported systems and hit the latent to furnish far majestic show at substantially inferior production costs. These characteristics allow the development of an entirely new generation of standoff weapons, built around FPAs and high powered signal processing chips. These weapons will dispense hundreds of intelligent sub munitions, each with the ability to see, recognize and priorities targets self-sufficiently of the launch vehicle or mother projectile.

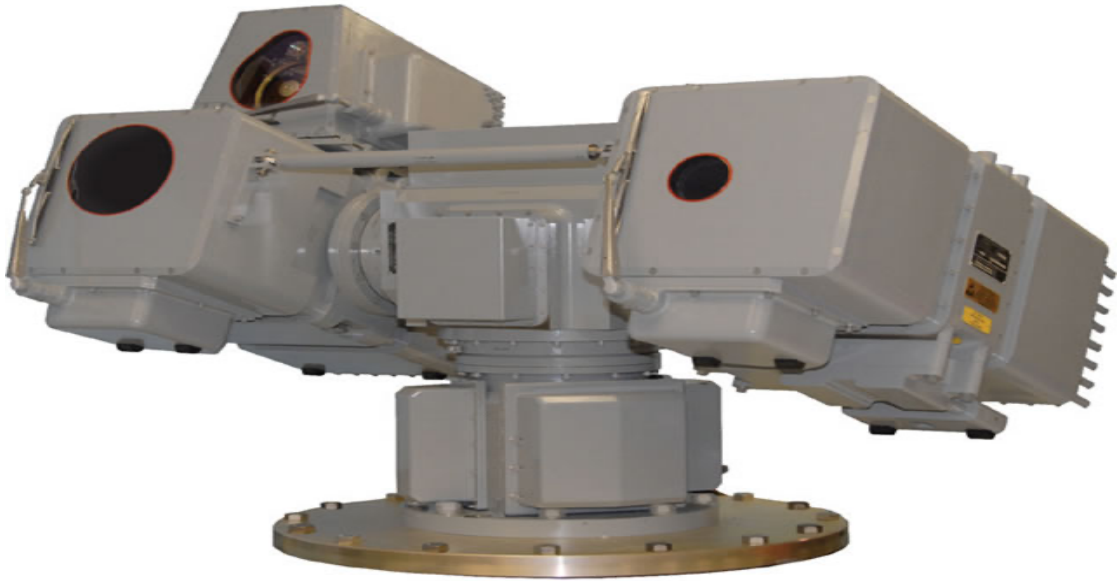

Figure 3.7: Electro-Optical Sensor (MK20_EOSS_L3)

Seeing these necessities, satellite sensors are settled which contain many discrete picture particles or pixels so that the system can detect and track the missiles within a very large area. In order to track the missiles to the preferred accuracy, it is very important that the pixels have a very high resolution so that they can observe a large area within few seconds. To do this, several millions of pixels are frequently needed. The detection system using this large pixel arrays are much heavy (about 6,000 to 8,000 pounds). Thus, it is highly necessary to find out a cost effective missile detection system according to the economic condition of Bangladesh. So, in that context, Our thesis work is based on finding out a cost effective solution that can be implemented in our country.

## 3.6 Radar Frequency Bands

Radar systems transmit electromagnetic, or radio waves. The reflected radio waves can be detected by the radar system receiver. The frequency of the radio waves used depends upon the radar application and are commonly classified according to Table 1.

The required antenna size is proportional to wavelength and therefore inversely proportional to frequency. The ability of the radar to focus the radiated and received energy in a narrow region is also dependent upon both antenna size and frequency choice.

A larger antenna allows the beam to be more tightly focused at a given frequency. The "focusing" ability of the antenna is often described using an antenna lobe diagram (see Figure 1), which plots the directional gain of an antenna over the azimuth (side to side) and elevation (up and down).

Most airborne radars operate between the L and Ka bands, also known as the microware region. Many short range targeting radars, such as on a tank or helicopter, operate in the millimeter band. Many long range ground based operate at UHF or lower frequencies, due to the ability to use large antennas and minimal atmospheric attenuation and ambient noise.

| Radar Band | Frequency (GHz) | Wavelength (cm) |
|---|---|---|
| Millimeter | 40 to 100 | 0.75 to 0.30 |
| Ka | 26.5 to 40 | 1.1 to 0.75 |
| K | 18 to 26.5 | 1.7 to 1.1 |
| Ku | 12.5 to 18 | 2.4 to 1.7 |
| X | 8 to 12.5 | 3.75 to 2.4 |
| C | 4 to 8 | 7.5 to 3.75 |
| S | 2 to 4 | 15 to 7.5 |
| L | 1 to 2 | 30 to 15 |
| UHF | 0.3 to 1 | 100 to 30 |

Table 3.1 : Radar bands, frequencies, and wavelengths

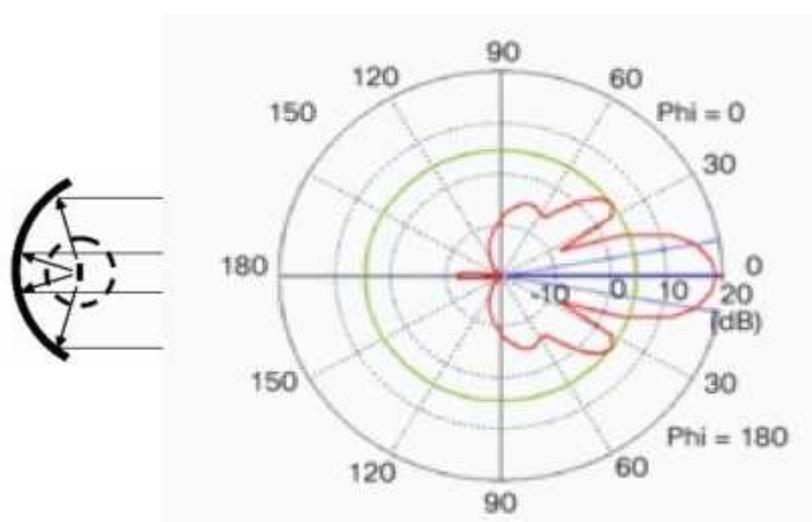

Figure 3.8: Antenna gain plot.

## 3.7 Radar Range Equations

Detection of objects using radar involves sophisticated signal processing. For a single pulse the following signal to noise ratio is:

$$SNR = P_t \, G_t \, A_r \, \sigma \, t_{pulse} / ((4\pi)^2 \, R^4 \, k \, T_s \, L)$$

Where:
- $P_t$ = transmitted power
- $G_t$ = antenna transmit gain
- $A_r$ = Receive antenna aperture area
- $\sigma$ = target radar cross section (function of target geometric cross section, reflectivity of surface, and directivity of reflections)
- $t_{pulse}$ = duration of receive pulse
- $L$ = aggregate of system losses
- $R$ = range between radar and target
- $k$ = boltsmann's constant
- $T_s$ = system noise temperature

Observe that the received power drops with the fourth power of the range, so radar systems must cope with very large dynamic ranges in the receive signal processing. The radar energy seen by the target drops proportional to the range squared. The reflected energy seen by the radar receiver further drops by a factor of the range squared. The ability to detect very small signals in the presence of large interfering signals is crucial to operate at longer ranges.

# CHAPTER 4

# MISSILE TRACKING SYSTEM

## 4.1 Introduction

The dominance of radar seekers in missile guidance at the terminal phase has been well well-known since 1960s. Three basic configurations, viz., semi active, active, and passive have been executed in various types of seekers, including the track via-missile concept used in Patriot missiles during Gulf War I. However, active radar seekers are the most popular in all the current missile programs owing to their flexibility of design and implementation to suit almost every mission requirement apart from all-weather capability. This is primarily due to the choice of waveform design, optimization of receiver, and adaptability and flexibility offered by the digital signal processing techniques in vogue. The most extensively employed configuration of active radar seekers so far realized is the coherent mono pulse tracker with a gimbaled antenna structure.

In our work, an attempt has been made to highlight the features of conventional active radar seekers and the associated current technologies, followed by the requirements of our desired technology.

## 4.2 Active Radar Seeker

The active radar seeker, from a radar engineer's view, may be defined as an application-specific solid missile-borne mono pulse tracking radar whose antenna is mounted on a gyro-stabilized platform such that the antenna is isolated /decoupled from the body movement of the missile. The above basic idea stems from the requirement of generating highly accurate target information necessary for precise homing guidance of the missile.

The following figure shows the block diagram of an active radar seeker:

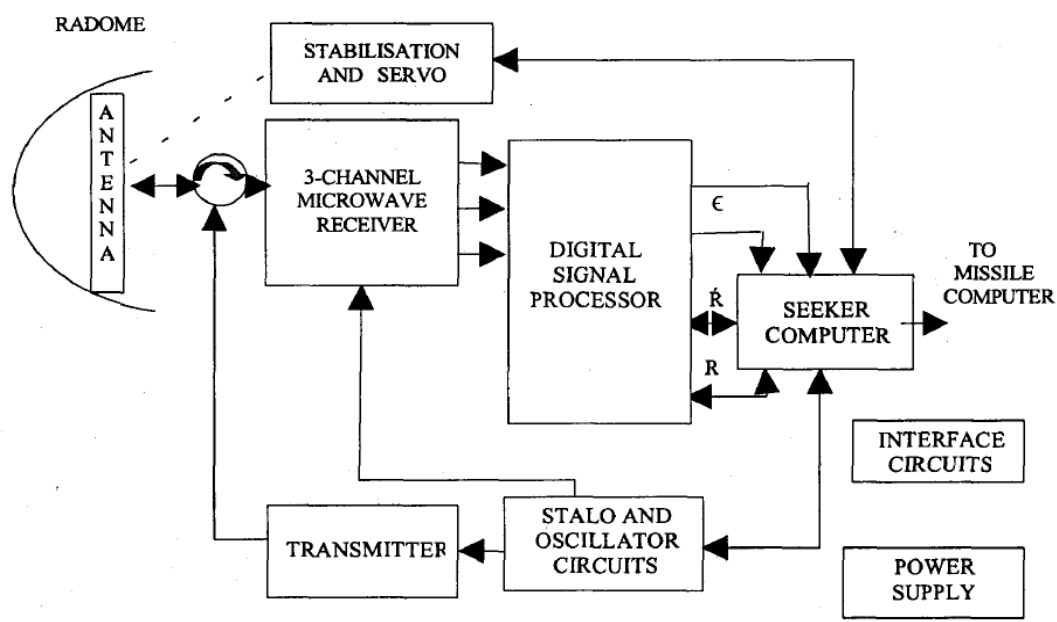

Figure 4.1: Active Radar Seeker (Basic Blocks)

The active radar seeker basically comprises of blocks, shown in Fig. 3.1, configured as a coherent three-channel mono pulse master oscillator power amplifier (MOPA) system capable of tracking the target in terms of angle as well as relative velocity (in terms of Doppler frequency shift, A) and range, R (optional). To accomplish homing guidance, necessary guidance commands must be computed by the missile computer known as onboard computer (OBC), which requires the measured value of rate of change of line-of-sight to the target with respect to the missile. By keeping a continuous track of the target in terms of angle by the mono pulse tracker such line-of-sight rate is computed which is self-explanatory. Additional target information in terms of range and/or relative velocity (Doppler frequency shift) is used to identify and track a desired target among the other targets including clutter, if any. In Fig. 3.2, the operation of the active radar seeker has been shown in relation to the onboard computer responsible for overall guidance of the missile.

The radar sensor, for most of the surface-to air missiles (SAMs) as well as air-to-air missiles (AAMs), is configured as a high PRF (HPRF) pulsed (Doppler) radar frequency essentially as a Doppler tracker apart from being a basic mono pulse angle tracker. Frequency of operation varies from X-band to Ku-band for SAMs and AAMs to millimeter wave frequencies (35 GHz and 94 GHz) for SAMs in air defense role. Seekers for precision-guided munitions (PGMs) and antitank missiles (ATMs) also operate at these millimeter wave frequencies. However, it may be noted that for anti-ship missiles (ASMs), ATMs and PGMs, the radar waveform is for seeker stabilization (strapped-down and identification. configuration).

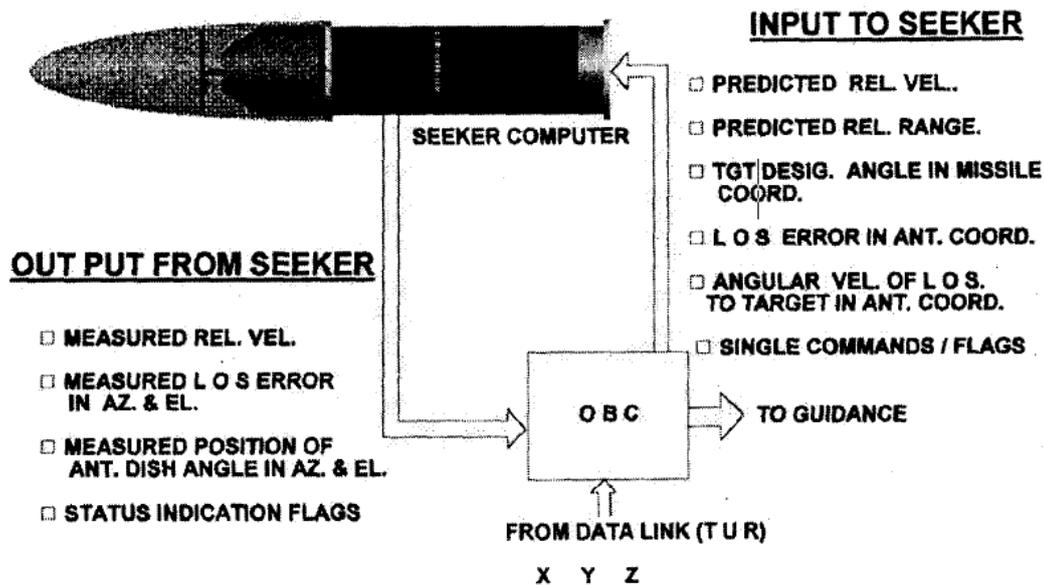

Figure 4.2: Active radar seeker operation with onboard computer

## 4.3 Microcontroller Based Missile Tracking System

This project is being implemented with multi sensors. By getting inputs from all the sensors, we will be able to be sure about the object. As sensors always follow the principle of logical operations, if the system works with sonar sensor, smoke detection sensor and IR sensor then it must be able to detect the object perfectly. The code for the missile tracking system is attached with the paper. The components used in the system are described below:

### 4.3.1 Sonar Sensor

Sonar (originally an acronym for Sound Navigation and Ranging) is a system that uses sound propagation to navigate, communicate with or detect objects on or under the surface of the water, such as other vessels. Two types of technology share the name "sonar": passive sonar is basically listening for the sound made by vessels; active sonar is emitting pulses of sounds and listening for echoes. Sonar may be used as a means of acoustic location and of measurement of the echo characteristics of "targets" in the water. Acoustic location in air was used before the introduction of radar. Sonar may also be used in air for robot navigation, and SODAR (upward looking in-air sonar) is used for atmospheric research. The term sonar is also used for the equipment used to generate and receive the sound. The acoustic frequencies used in sonar systems vary from very low (infrasonic) to extremely high (ultrasonic). The study of underwater sound is known as underwater acoustics or hydro acoustics.

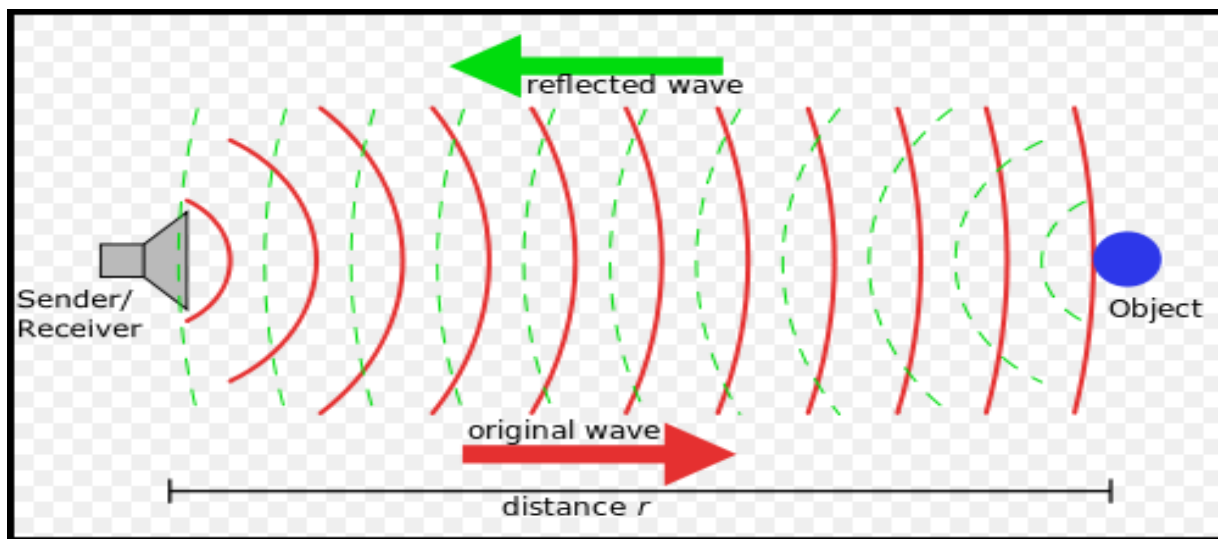

Figure 4.3: working principle of SONAR

Active sonar uses a sound transmitter and a receiver. When the two are in the same place it is mono-static operation. When the transmitter and receiver are separated it is bi-static operation. When more transmitters (or more receivers) are used, again spatially separated, it is multi-static operation. Most sonar is used mono-statically with the same array often being used for transmission and reception. Active son buoy fields may be operated multi-statically.

Active sonar creates a pulse of sound, often called a "ping", and then listens for reflections (echo) of the pulse. This pulse of sound is generally created electronically using a sonar projector consisting of a signal generator, power amplifier and electro-acoustic transducer/array. A beam former is usually employed to concentrate the acoustic power into a beam, which may be swept to cover the required search angles. Generally, the electro-acoustic transducers are of the Tonpilz type and their design may be optimized to achieve maximum efficiency over the widest bandwidth, in order to optimize performance of the overall system. Occasionally, the acoustic pulse may be created by other means, e.g. (1) chemically using explosives, or (2) air guns or (3) plasma sound sources.

To measure the distance to an object, the time from transmission of a pulse to reception is measured and converted into a range by knowing the speed of sound. To measure the bearing, several hydrophones are used, and the set measures the relative arrival time to each or with an array of hydrophones, by measuring the relative amplitude in beams formed through a process called beam forming. Use of an array decreases the spatial response so that to provide wide cover multi-beam systems are used. The target signal (if present) together with noise is then passed through various forms of signal dispensation, which for simple sonars may be just energy measurement. It is then presented to some form of decision device that calls the output either the necessary signal or noise. This decision device may be an operator with headphones or a display, or in more classy sonars this function may be carried out by software. Further processes may be carried out to classify the target and localize it, as well as calculating its velocity.

The pulse may be at perpetual frequency or a chirp of changing frequency (to allow pulse compression on reception). Simple sonars generally use the former with a filter wide enough to cover possible Doppler changes due to target movement, while more complex ones generally include the latter technique. Since digital processing became available pulse compression has usually been implemented using digital correlation techniques. Military sonars often have multiple beams to provide all-round cover while simple ones only cover a narrow arc, although the beam may be rotated, relatively slowly, by mechanical scanning.

Mainly when single frequency transmissions are used, the Doppler effect can be used to measure the radial speed of a target. The variance in frequency between the transmitted and received signal is measured and converted into a velocity. Since Doppler shifts can be introduced by either receiver or target motion, allowance has to be made for the radial speed of the searching platform.

One useful small sonar is similar in appearance to a waterproof flashlight. The head is pointed into the water, a button is pressed, and the device displays the distance to the target. Another variant is a "fish finder" that shows a small display with shoals of fish. Some civilian sonars (which are not designed for stealth) approach active military sonars in capability, with quite exotic three-dimensional displays of the area near the boat.

When active sonar is used to measure the distance from the transducer to the bottom, it is known as echo sounding. Similar methods may be used viewing upward for wave measurement.

Active sonar is also used to measure distance through water between two sonar transducers or a combination of a hydrophone (underwater acoustic microphone) and projector (underwater acoustic speaker). A transducer is a device that can transmit and receive acoustic signals ("pings"). When a hydrophone/transducer receives a specific interrogation signal it responds by transmitting a specific reply signal. To measure distance, one transducer/projector transmits an interrogation signal and measures the time between this transmission and the receipt of the other transducer/hydrophone reply. The time difference, scaled by the speed of sound through water and divided by two, is the distance between the two platforms. This technique, when used with multiple transducers/hydrophones/projectors, can calculate the relative positions of static and moving objects in water.

In combat situations, an active pulse can be detected by an opponent and will reveal a submarine's position.

A very directional, but low-efficiency, type of sonar (used by fisheries, military, and for port security) makes use of a complex nonlinear feature of water known as non-linear sonar, the virtual transducer being known as a parametric array.

### 4.3.2 Smoke Sensor

A smoke detector is a device that senses smoke, typically as an indicator of fire. Smoke can be detected either optically (photoelectric) or by physical process (ionization); detectors may use either, or both, methods. Sensitive alarms can be used to detect, and thus deter, smoking in areas. A photoelectric, or optical smoke detector contains a source of infrared, visible, or ultraviolet light (typically an incandescent light bulb or light-emitting diode), a lens, and a photoelectric receiver (typically a photodiode). In spot-type detectors all of these components are arranged inside a chamber where air, which may contain smoke from a nearby fire, flows.

In large open areas such as atria and auditoriums, optical beam or projected-beam smoke detectors are used instead of a chamber within the unit: a wall-mounted unit emits a beam of infrared or ultraviolet light which is either received and processed by a separate device, or reflected back to the receiver by a reflector[10].

In some types, particularly optical beam types, the light emitted by the light source passes through the air being tested and reaches the photo sensor. The received light intensity will be reduced due to scattering from particulates of smoke, air-borne dust, or other substances; the circuitry detects the light intensity and generates the alarm if it is below a specified threshold, potentially due to smoke. In other types, typically chamber types, the light is not directed at the sensor, which is not illuminated in the absence of particles. If the air in the chamber contains particles (smoke or dust), the light is scattered and some of it reaches the sensor, triggering the alarm.[11]

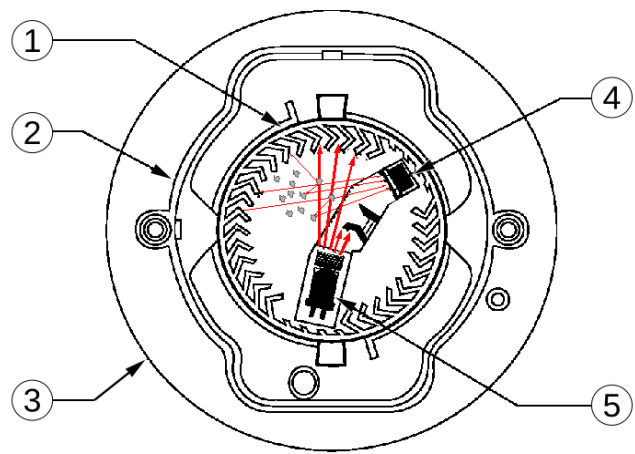

Figure 4.4: Smoke (Optical) Sensor

Optical smoke detector

1: Optical chamber

2: Cover

3: Case molding

4: Photodiode (transducer)

5: Infrared LED

In this paper, we will use the sensor analog output voltage and when the smoke reaches a certain level, it will make sound a buzzer and a red LED will turn on. When the output voltage is below that level, a green LED will be on.

### 4.3.2.1 What is an MQ-2 Smoke Sensor?

The MQ-2 smoke sensor is sensitive to smoke and to the following flammable gases:

- LPG
- Butane
- Propane
- Methane
- Alcohol
- Hydrogen

The resistance of the sensor is different depending on the type of the gas.

The smoke sensor has a built-in potentiometer that allows you to adjust the sensor sensitivity according to how accurate you want to detect gas.

### 4.3.2.2 How does it Work?

The voltage that the sensor outputs changes accordingly to the smoke/gas level that exists in the atmosphere. The sensor outputs a voltage that is proportional to the concentration of smoke/gas.

In other words, the relationship between voltage and gas concentration is the following:

- The greater the gas concentration ,the greater the output voltage
- The lower the gas concentration ,the lower the output voltage

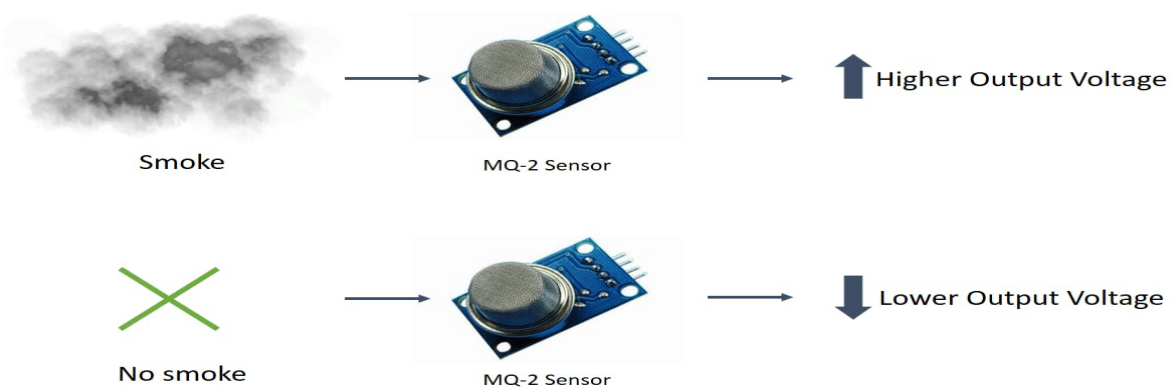

Figure 4.5: Working principle of Smoke Sensor

The output can be an analog signal (A0) that can be read with an analog input of the Arduino or a digital output (D0) that can be read with a digital input of the Arduino.

### 4.3.3 Infrared Sensor and it's Homing

A passive infrared sensor (PIR sensor) is an electronic sensor that measures infrared (IR) light radiating from objects in its field of view. They are most often used in PIR-based motion detectors. All objects with a temperature above absolute zero emit heat energy in the form of radiation. Usually this radiation isn't visible to the human eye because it radiates at infrared wavelengths, but it can be detected by electronic devices designed for such a purpose.

The term *passive* in this instance refers to the fact that PIR devices do not generate or radiate energy for detection purposes. They work entirely by detecting infrared radiation emitted by or reflected from objects. They do not detect or measure "heat" [15][16].

An individual PIR sensor detects changes in the amount of infrared radiation impinging upon it, which varies depending on the temperature and surface characteristics of the objects in front of the sensor. When an object, such as a human, passes in front of the background, such as a wall, the temperature at that point in the sensor's field of view will rise from room temperature to body temperature, and then back again. The sensor converts the resulting change in the incoming infrared radiation into a change in the output voltage, and this triggers the detection. Objects of similar temperature but different surface characteristics may also have a different infrared emission pattern, and thus moving them with respect to the background may trigger the detector as well [16].

PIRs come in many configurations for a wide variety of applications. The most common models have numerous Fresnel lenses or mirror segments, an effective range of about ten meters (thirty feet), and a field of view less than 180 degrees. Models with wider fields of view, including 360 degrees, are available—typically designed to mount on a ceiling. Some larger PIRs are made with single segment mirrors and can sense changes in infrared energy over thirty meters (one hundred feet) away from the PIR. There are also PIRs designed with reversible orientation mirrors which allow either broad coverage (110° wide) or very narrow "curtain" coverage , or with individually selectable segments to "shape" the coverage.

Infrared homing is a passive weapon guidance system which uses the infrared (IR) light emission from a target to track and follow it. Missiles which use infrared seeking are often referred to as "heat-seekers", since infrared is radiated strongly by hot bodies. Many objects such as people, vehicle engines and aircraft generate and emit heat, and as such, are especially visible in the infrared wavelengths of light compared to objects in the background.

Infrared seekers are passive devices, which, unlike radar, provide no indication that they are tracking a target. This makes them suitable for sneak attacks during visual encounters, or over longer ranges when used with a forward looking infrared system or similar cuing system. This makes heat-seekers extremely deadly; 90% of all United States air combat losses over the past 25 years have been due to infrared-homing missiles. They are, however, subject to a number of simple countermeasures, most notably dropping flares behind the target to provide false heat sources. This only works if the pilot is aware of the missile and deploys the countermeasures, and modern seekers have rendered these increasingly ineffective even in that case.

The first IR devices were experimented with in the pre-World War II era. During the war, German engineers were working on heat seeking missiles and proximity fuses, but did not have time to complete development before the war ended. Truly practical designs did not become possible until the introduction of conical scanning and miniaturized vacuum tubes during the war. Anti-aircraft IR systems began in earnest in the late 1940s, but both the electronics and entire field of rocketry was so new that it required considerable development before the first examples entered service in the mid-1950s. These early examples had significant limitations and achieved very low success rates in combat during the 1960s. A new generation developed in the 1970s and 80s made great strides and significantly improved their lethality. The latest examples from the 1990s and on have the ability to attack targets out of their field of view (FOV), behind them, and even pick out vehicles on the ground .The infrared sensor package on the tip or head of a heat-seeking missile is known as the *seeker head*. [13]

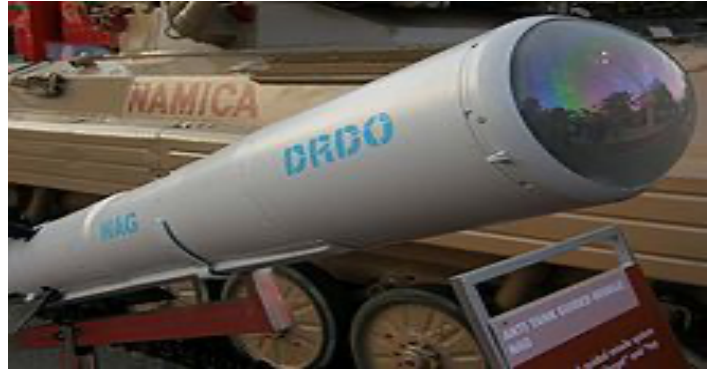

Figure 4.6: Working principle of Smoke Sensor

Most infrared guided missiles have their seekers mounted on a gimbal. This allows the sensor to be pointed at the target when the missile is not. This is important for two main reasons. One is that before and during launch, the missile cannot always be pointed at the target. Rather, the pilot or operator points the seeker at the target using radar, a helmet-mounted sight, an optical sight or possibly by pointing the nose of the aircraft or missile launcher directly at the target. Once the seeker sees and recognizes the target, it indicates this to the operator who then typically "uncages" the seeker (which is allowed to follow the target). After this point the seeker remains locked on the target, even if the aircraft or launching platform moves. When the weapon is launched, it may not be able to control the direction it points until the motor fires and it reaches a high enough speed for its fins to control its direction of travel. Until then, the gimbaled seeker needs to be able to track the target independently.

Finally, even while it is under positive control and on its way to intercept the target, it probably will not be pointing directly at it; unless the target is moving directly toward or away from the launching platform, the shortest path to intercept the target will not be the path taken while pointing straight at it, since it is moving laterally with respect to the missile's view. The original heat-seeking missiles would simply point towards the target and chase it; this was inefficient. Newer missiles are smarter and use the gimbaled seeker head combined with what is known as proportional guidance in order to avoid oscillation and to fly an efficient intercept path.

## 4.3.4 Infrared Countermeasure

An infrared countermeasure (IRCM) is a device designed to protect aircraft from infrared homing ("heat seeking") missiles by confusing the missiles' infrared guidance system so that they miss their target (electronic countermeasure). Heat-seeking missiles were responsible for about 80% of air losses. The most common method of infrared countermeasure is deploying flares, as the heat produced by the flares creates hundreds of targets for the missile.

An infrared sensor that is sensitive to heat, such as emitted from an aircraft engine, is included on missiles launched by man-portable air-defense systems (MANPAD). Using a steering system, the missile is programmed to home in on the infrared heat signal. Because they are portable, MANPAD missiles have a limited ranged, burning out a few seconds after launch. Because they are expensive, such countermeasure systems have been seldom used, primarily on military aircraft.

Infrared missile seekers of the first generation typically used a spinning reticle with a pattern on it that modulates infrared energy before it falls on a detector (A mode of operation called Spin scan). The patterns used differ from seeker to seeker, but the principle is the same. By modulating the signal, the steering logic can tell where the infrared source of energy is relative to the missile direction of flight. In more recent designs the missile optics will rotate and the rotating image is projected on a stationary reticle (a mode called Conical scan) or stationary set of detectors which generates a pulsed signal which is processed by the tracking logic. Most shoulder-launched (MANPADS) systems use this type of seeker, as do many air defense systems and air-to-air missiles (for example the AIM-9L).

Countermeasure systems are usually integrated into the aircraft, such as in the fuselage, wing, or nose of the aircraft, or fixed to an outer portion of the aircraft. Depending on where the systems are mounted, they can increase drag, reducing flight performance and increasing operating cost. Much time and money is spent on testing, maintaining, servicing, and upgrading systems. These procedures require that the aircraft are grounded for a period of time.

Infrared seekers are designed to track a strong source of infrared radiation (usually a jet engine in modern military aircraft). IRCM systems are based on a source of infrared radiation with a higher intensity than the target. When this is received by a missile, it may overwhelm the original infrared signal from the aircraft and provide incorrect steering cues to the missile. The missile may then deviate from the target, breaking lock. Once an infrared seeker breaks lock (they typically have a field of view of 1 - 2 degrees), they rarely reacquire the target. By using flares, the target can cause the confused seeker to lock onto a new infrared source that is rapidly moving away from the true target.

The modulated radiation from the IRCM generates a false tracking command in the seeker tracking logic. The effectiveness of the IRCM is determined by the ratio of jamming intensity to the target (or signal) intensity. This ratio is usually called the J/S ratio. Another important factor is the modulation frequencies which should be close to the actual missile frequencies. For spin scan missiles the required J/S is quite low but for newer missiles the required J/S is quite high requiring a directional source of radiation[12].

### 4.3.5 Infrared Search and Track

An infrared search and track (IRST) system (sometimes known as infrared sighting and tracking) is a method for detecting and tracking objects which give off infrared radiation such as jet aircraft and helicopters[15].

IRST is a generalized case of forward looking infrared (FLIR), i.e. from forward-looking to all-round situation awareness. Such systems are passive (thermo graphic camera), meaning they do not give out any radiation of their own, unlike radar. This gives them the advantage that they are difficult to detect.

However, because the atmosphere attenuates infrared to some extent (although not as much as visible light) and because adverse weather can attenuate it also (again, not as badly as visible systems), the range compared to a radar is limited. Within range, angular resolution is better than radar due to the shorter wavelength.

Detection range varies with

- clouds
- altitude
- air temperature
- target's attitude
- target's speed

The higher the altitude, the less dense the atmosphere and the less infrared radiation it absorbs - especially at longer wavelengths. The effect of reduction in friction between air and aircraft does not compensate the better transmission of infrared radiation. Therefore, infrared detection ranges are longer at high altitudes.

At high altitudes, temperatures range from −30 to −50 °C - which provide better contrast between aircraft temperature and background temperature.

The Eurofighter Typhoon's PIRATE IRST can detect subsonic fighters from 50 km from front and 90 km from rear[14] - the larger value being the consequence of directly observing the engine exhaust, with an even greater increase being possible if the target uses afterburners.

The range at which a target can be sufficiently confidently identified to decide on weapon release is significantly inferior to the detection range - manufacturers have claimed it is about 65% of detection range.

### 4.3.6 Arduino UNO R3

There are various types of Arduino boards in which many of them were third-party compatible versions. The most official versions available are the Arduino Uno R3 and the Arduino Nano V3. Both of these run a 16MHz Atmel ATmega328P 8-bit microcontroller with 32KB of flash RAM 14 digital I/O and six analogue I/O and the 32KB will not sound like as if running Windows. Arduino projects can be stand-alone or they can communicate with software on running on a computer. For e.g. Flash, Processing, Max/MSP). The board is clocked by a 16 MHz ceramic resonator and has a USB connection for power and communication. You can easily add micro SD/SD card storage for bigger tasks.

Arduino uno microcontroller can sense the environment by receiving input from a variety of sensors and can affect its surroundings by controlling lights, motors, and other actuators. The microcontroller is programmed using the Arduino programming language (based on Wiring) and the Arduino development environment (based on Processing).

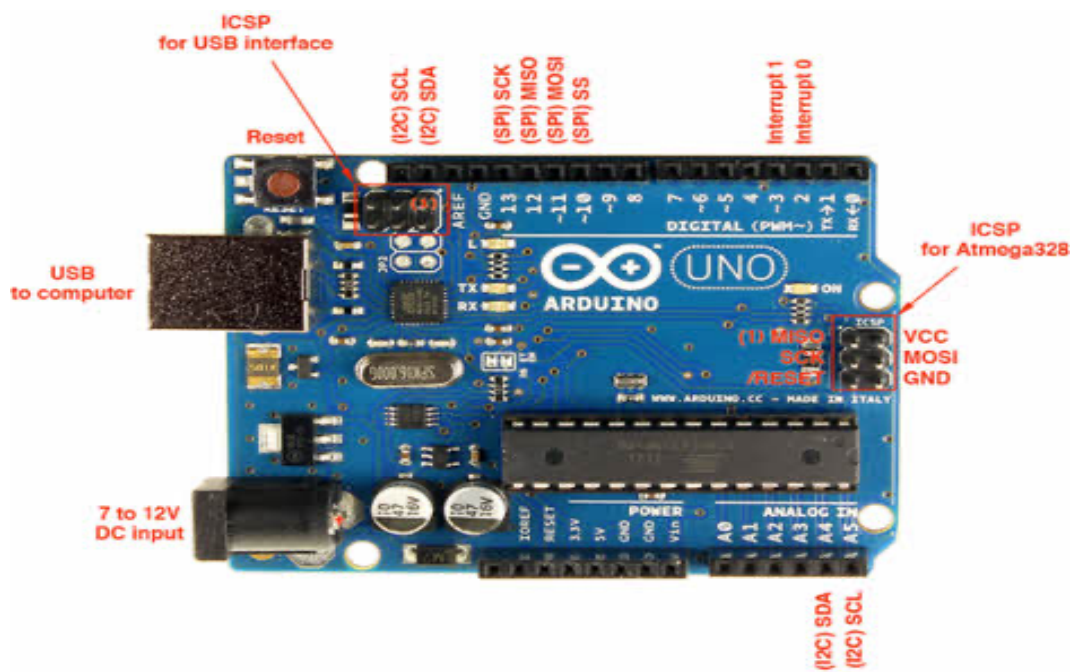

Figure 4.7: Arduino UNO R3

## 4.4 Block Diagram and Flowchart of the System

The Block Diagram and the flowchart of the microcontroller based system is shown in figure

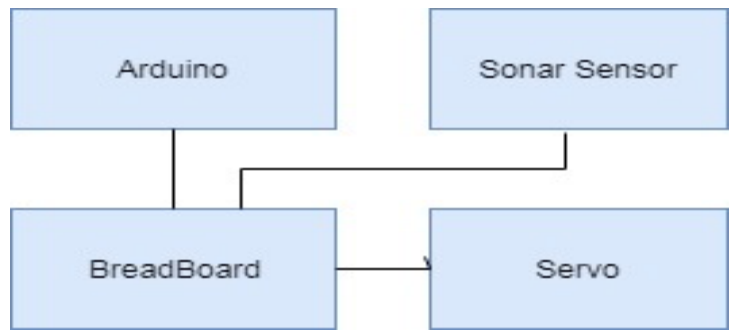

Figure 4.8: Block Diagram

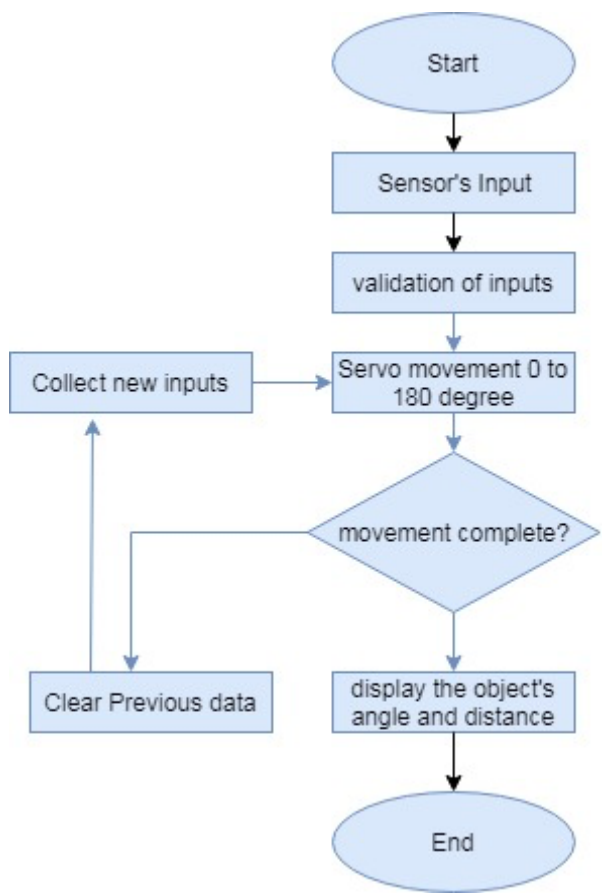

Figure 4.9: Flowchart

# CHAPTER 5

# TARGET ENGAGEMENT AND DESTRUCTION

## 5.1 Introduction

After the detection and the tracking phase of the missile, the next step is to engage the target with the anti-missile system and finally to destroy the desired missile. The elaborate description of the engagement and destruction technology is out of the scope of our research. But for the overall view of the system the engagement and destruction system will be discussed in brief. At first one computation algorithm will be given which will show how the overall system will work. Then few engagement systems will de described with necessary equations.

## 5.2 Computation Algorithm

The algorithm represents the effectiveness computation of armored target kill probability by the use of guided and unguided anti-armor rocket-projectiles, in cases when the firing is performed from any kind of motionless infantry antitank related weapons or mobile platforms (of aircraft or helicopter type) is a relatively complex one (Fig.3)

The algorithm is based on the data relative to: firing position, target, information on the weapon and rocket projectile in the combat system, gunner's qualities, method of preparing for firing and the firing itself, and finally the prevailing situation in the field.

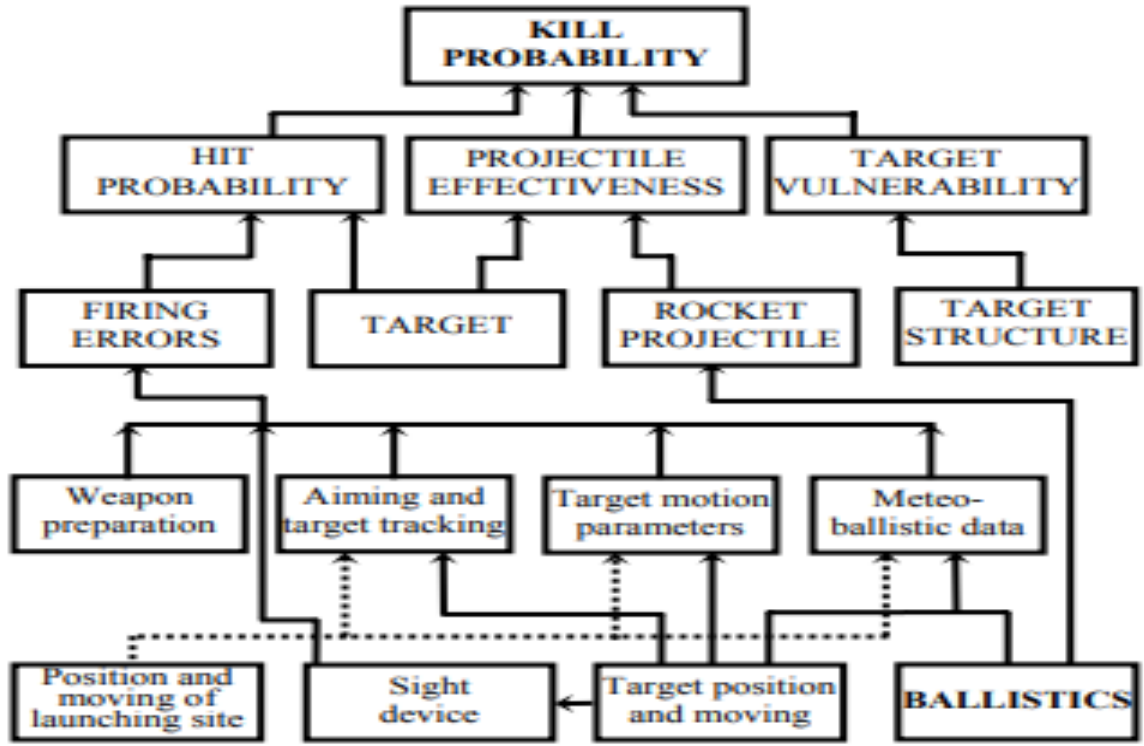

Fig 5.1: Effectiveness Computation Algorithm

## 5.3  Undefended Target - One-Sided Engagements

Any kind of Basic missile system exchange models always follow the probability theory. A target destroying probability of a single missile (3). This quantity is known as the single shot probability of kill and is formally denoted by PKSS (probability of kill single shot). Sequential exchanges of missiles between forces can be modeled to determine either the expected number of survivors or kills. As example, for a one-sided exchange between a fighter aircraft carrying air-to-air missiles (AAM) and a bomber not carrying AAM, the equations are simple. Equation 1 represents the probability of the bomber surviving the fighter attack and equation 2 represents the probability of the fighter destroying the target.

$$S(k) = (1-p)^k \quad (1)$$

$$P(k) = 1 - S(k) \quad (2)$$

k -is the number of shots

p -is the probability of a single shot destroying the target

S(k) -is the probability of target surviving salvo of shots

P(k) -is the probability of target destroyed by salvo of shots

Figure 1 illustrates equation 1 and 2 for different values of p and k. Although, this model is extremely simple, it is appropriate for modeling the attack of a slow-moving undefended target.

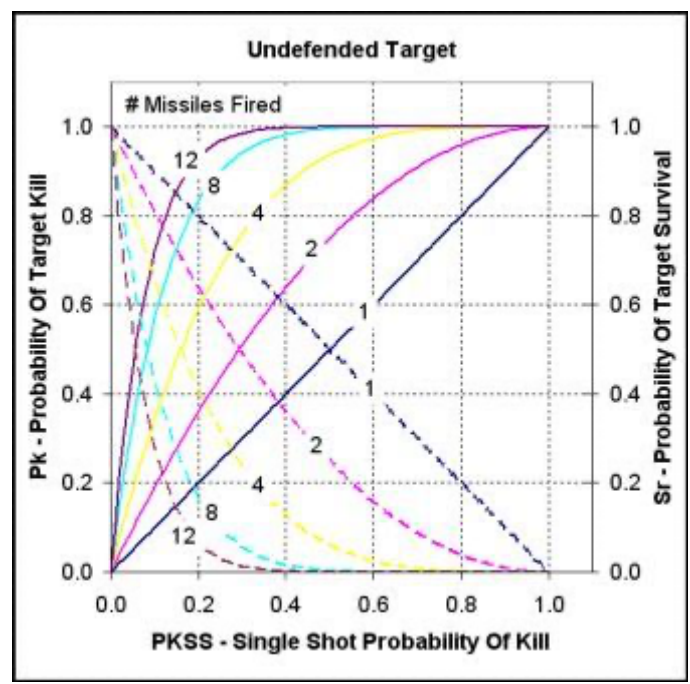

Figure 5.2: Single Shot Probability of Kill

## 5.4   One Versus One Sequential Engagements

A similar model can be developed for an exchange of missiles between two aircraft (3). This is known as a 1v1 engagement where v stands for versus. As is convention, one side will be known as blue and the other red. For example, we will assume blue fires first. Blue fires first and the probability of red being destroyed is pB and the probability that red survives is 1-pB. Red responds and the probability of blue being destroyed is now pR(1-pB) and probability of blue surviving is 1-pR(1-pB). The sequential exchange continues until both sides are out of weapons. When blue shoots first, equation 3 and 4 are used to represent the probability of red being destroyed by blue after n blue weapons are fired and the probability of blue being destroyed by n red weapons, respectively. Conversely if red fires first, equation 5 and 6 are used. Figure 2a illustrates equations 3 and 4 for various values of pB and pR, and figure 2b illustrates equation 5 and 6.

$$P_{BR}(n) = \frac{p_B\left[1-(1-p_B)^n(1-p_R)^n\right]}{1-(1-p_B)(1-p_R)} \quad (3)$$

$$P_{BB}(n) = \frac{p_R(1-p_B)\left[1-(1-p_B)^n(1-p_R)^n\right]}{1-(1-p_B)(1-p_R)} \quad (4)$$

$$P_{RR}(n) = \frac{p_R(1-p_B)\left[1-(1-p_B)^n(1-p_R)^n\right]}{1-(1-p_B)(1-p_R)} \quad (5)$$

$$P_{RB}(n) = \frac{p_R\left[1-(1-p_B)^n(1-p_R)^n\right]}{1-(1-p_B)(1-p_R)} \quad (6)$$

PBR(n) - when blue shoots first, probability red is destroyed
PBB(n) - when blue shoots first, probability blue is destroyed
PRR(n) - when red shoots first, probability red is destroyed
PRB(n) - when red shoots first, probability blue is destroyed
pB - blue probability of single shot destroying red
pR - red probability of single shot destroying blue
n - number of missiles fired by red/blue

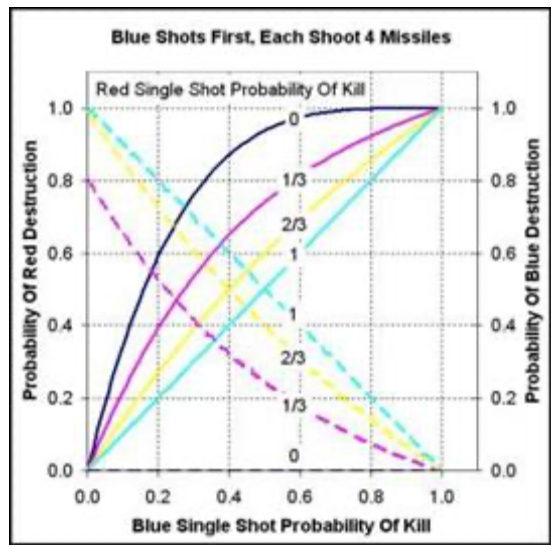

Figure 5.2(a)

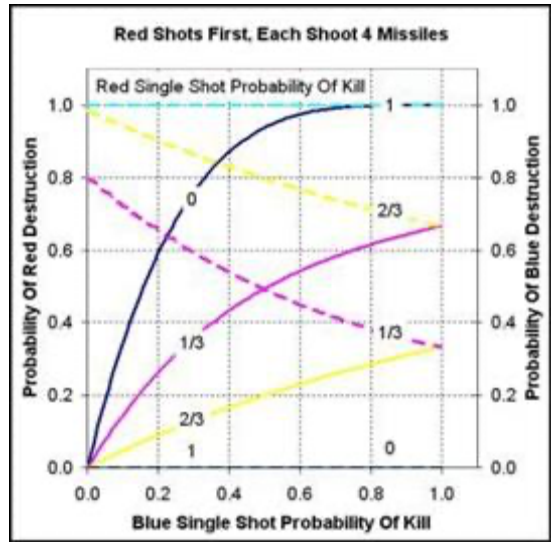

Figure 5.2(b)

The obvious conclusion from figure 2a and 2b is shooting first is very important, even more important than weapon PKSS.

## 5.5 Many Versus Many One-Sided Engagements

The model in section 5.1 can be extended beyond 1v1 engagements to 2v2, 2v4, or NvM in general (4, 5). The convention for the nomenclature is the number of blue aircraft is listed first, then v, and finally the number of red aircraft.

Three cases have been developed in the reference 4 for dealing with target sorting. In one extreme, each attacker randomly picks its targets. In this case, some targets may be double targeted and weapons may be wasted. This case may represent when there is poor command-and-control and/or poor communication between the individual attackers. The other extreme is perfect target sorting where all attackers uniformly distribute weapons over the targets. This case represents when the attackers and/or their controllers have good communication and the ability to effectively sort targets. The third case is an extension of the second where the attackers and/or their controllers have the ability to make a good kill assessment allowing them to focus new salvos only on surviving targets.

This paper will focus on the simple extension of the 1v1 engagement to an NvN engagement. It is important to demonstrate that 1v1 results can be extended to many-on-many, but the details can be worked out in latter. In this situation N = M, and each attacker is assigned a single target. In addition the attacker expends all its missiles on its assigned target. Equations 7-10 can be used to determine the average number of survivors after the engagement is complete.

$$N = MJ \quad (7)$$

$$Sr = (1-p_J)^{N/T} \quad (8)$$

$$p_J = 1 - (1-p)^J \quad (9)$$

$$E_s = T \cdot Sr \quad (10)$$

J - number of weapons per attacker
M - number of attackers
N - total number of weapons
T - number of targets
$p_J$ - probability of kill for J shots
p - probability of a single shot destroying the target
Sr - average survivability of each target
$E_s$ - expected number of surviving targets

Since M = T, equations 7-10 can be simplified to equation 11.

$$Sr = (1-p)^{J \cdot J} \quad (11)$$

## 5.6 One Versus One Sequential Engagement - Mutual Kills

Although the chance of both sides getting destroyed simultaneously may seem remote, with missiles that are actively guided in the terminal phase it can occur. This type of missile only needs to be guided by the attacker to within a certain range of the target, and then the attacker may disengage from the fight. This is advantage compared to the less sophisticated semi-active guided missile that must be guided all the way to the target (4). The author was not able to find the derivation of a 1v1 sequential engagement in which both sides' shots arrive simultaneously resulting in a mutual kill. Therefore equations are derived for this situation. The missiles in this case no longer arrive sequentially, but each side's volley arrives simultaneously. With each volley there are four potential outcomes: both sides are destroyed, red is destroyed and blue survives, red survives and blue is destroyed, or both sides survive. For example, equation 12 is the probability red is destroyed in the first exchange. It is the sum of the probability both sides are destroyed, and red is destroyed and blue survives.

$$P_{1R} = p_B \cdot p_R + p_B(1 - p_R) \quad (12)$$

Similarly, equation 13 represents the probability red is destroyed in the second volley. It is the probability that red and blue survived the first volley multiplied by the two cases where red is destroyed.

$$P_{2R} = (1 - p_B)(1 - p_R)[p_B \cdot p_R + p_B(1 - p_R)] \quad (13)$$

The natural result of this procedure is a general equation for n shots. Equation 14 and 15 represent the probability of destruction after n shots for red and blue, respectively.

$$P_R = \frac{p_B(1 - (1 - p_B)^n (1 - p_R)^n)}{1 - (1 - p_B)(1 - p_R)} \quad (14)$$

$$P_B = \frac{p_R(1 - (1 - p_B)^n (1 - p_R)^n)}{1 - (1 - p_B)(1 - p_R)} \quad (15)$$

A careful reader will note the resulting equation 12 is the same as equation 3. Similarly equation 13 is the same as equation 6.

Figure 5.3 illustrates equation 12 and 13 for various values of pB and pR. As might be expected, the results fall somewhere between blue shoots first and red shoots first.

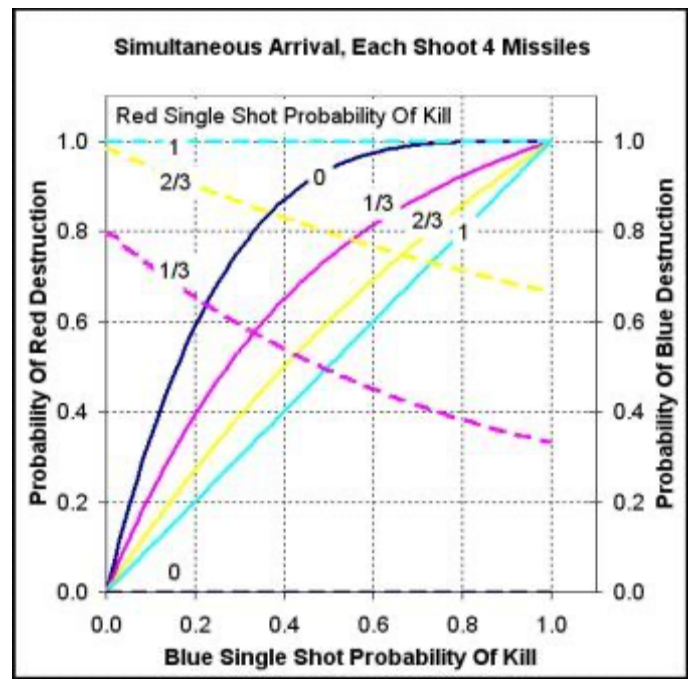

Figure 5.3

## 5.7 Including Engagement Geometry, Detection, Speed, And Different Weapon Technologies

The analytical models presented so far lack several important factors to modern air combat. For example there is no consideration of range and speed. They cannot answer whether it is feasible for each side to launch some number of weapons at each other or predict which side might get the first shot based on weapon system characteristics. To answer these deficiencies it is necessary to introduce ranges and speeds. Reference 5 develops an analytical model that includes these important factors. The model that follows, builds on that developed in reference 5 for our purposes.

We make the simplifying assumption that the engagement takes place in a plane. We develop the situation from the perspective that blue is the aggressor, but the model can be developed for either side as the aggressor. Blue desires to intercept red. Assuming a solution is possible, a collision course provides the minimum time to intercept. Frame T0 of figure 4 illustrates this situation in two dimensions. The line that intersects both aircraft is known as the line of sight or LOS. The heading of blue relative to the LOS is computed using equation 16 which is an application of the law of sine. Equation 17 can be used to find the closure rate which is an application of the law of cosine.

$$\beta = \arcsin\left(\frac{V_R}{V_B}\sin(\rho)\right) \quad \text{for} \quad \left|\frac{V_R}{V_B}\sin(\rho)\right| \leq 1 \quad (16)$$

$$V_C = V_B^2 + V_R^2 - 2 \cdot V_B \cdot V_R \cdot \cos(\alpha) \quad (17)$$

ρ - red heading relative to LOS

$V_R$ - red speed

$V_B$ - blue speed

β - blue heading relative to LOS

$V_C$ - closure rate

α - collision angle

Because the range of sin is [-1,1], arcsin will be undefined if $V_R/V_B *\sin(\rho)$ is outside this range. If $V_R/V_B < 1$ than this condition is met. If $V_R/V_B \geq 1$, but $\rho \leq \arcsin(V_B/V_R)$ then the condition is met. These are necessary conditions for a collision intercept to exist, but not sufficient. It is necessary and sufficient for a collision intercept if β exists and $V_C > 0$.

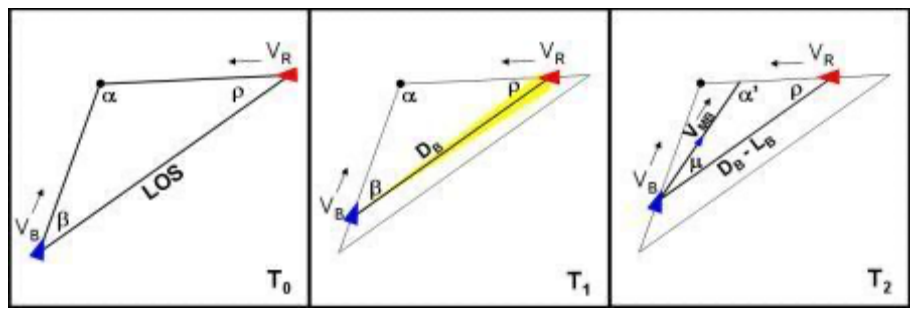

Figure 5.4

The probability of blue detecting red is assumed to be normally distributed with range. **$D_B$** is a random variable and represents the range at which blue detects red. There is also a normally distributed time delay before blue launches its missile at red after detection occurs. Equation 18 defines **$D_B$** and **$T_B$**. The time delay is represented by random variable **$T_B$**. The decrease in distance between blue and red during the time **$T_B$** is, of course, a function of **$T_B$** and therefore a random variable. It is represented by **L B** and given by equation 19. Therefore, the range at which blue launches its first missile at red is **D B - L B**.

$$D_B \sim N[d_B, \sigma_{DB}]$$
$$T_B \sim N[t_B, \sigma_{TB}] \quad (18)$$

$$L_B = V_C \cdot T_B$$
$$l_B = V_C \cdot t_B$$
$$\sigma_{LB} = V_C \cdot \sigma_{TB} \quad (19)$$

A desired characteristic of the model is the blue aircraft must detect the red aircraft before launching its missile. The normal distribution, assumed for simplicity, can violate this condition. Care must be taken to choose a $t_B$ and $\sigma_{TB}$ such that the probability of launching before detection is very small i.e. $t_B$ should be $3\sigma_{TB}$ away from 0.

The line the blue missile flies is different than the blue aircraft because the blue missile is traveling at a different speed. For simplicity it is assumed the speed at which the missile travels is the summation of the aircraft speed and a base missile speed. The line the missile follows is determined by equations 20 - 22.

$$\mu_B = \arcsin\left(\frac{V_R}{(V_{MB} + V_B)} \sin(\rho)\right) \quad (20)$$

$$V_{CMB}^2 = (V_{MB} + V_B)^2 + V_R^2 - 2 \cdot (V_{MB} + V_B) \cdot V_R \cdot \cos(\alpha_{MB}) \quad (21)$$

$$R_{MB} = \left(\frac{\sin(\rho)}{\sin(\alpha_{MB})}\right) \cdot (D_B - L_B) \quad (22)$$

V $_{MB}$ - blue missile speed without including speed imparted by launching aircraft
μ $_B$ - blue missile heading relative to line of sight
α $_{MB}$ - blue missile collision angle with red aircraft
V $_{CMB}$ - closure rate of blue missile and red aircraft
**R** $_{MB}$ - distance blue missile flies to get to red aircraft

The distance separating blue and red once the blue missile destroys red is **D B - L B - M B**. Equation 23 is used to calculate **M B**. The quantity **R** $_{MB}$ / V $_{CMB}$ is the time of flight (TOF) of the missile. Equation 24 is found by substituting for **M B** in **D B - L B - M B**

$$M_B = \frac{R_{MB}}{V_{CMB}} \cdot V_C \quad (23)$$

$$D_B - L_B - M_B = \left(1 - \frac{\sin(\rho)}{\sin(\alpha_{MB})} \cdot \frac{V_C}{V_{CMB}}\right) \cdot (D_B - L_B) \quad (24)$$

Radar seekers are most common for BVR missiles. There are two types of radar seekers used: semi-active radar (SAR) and active radar (AR). For air combat, AR seekers have been used on newer missiles. SAR guided missiles must be supported until impact with the target. AR guided missiles only have to be supported until the missile is within a certain range of the target (4). The advantages of this can be seen in the application of the above developed model.

As in reference 5, the probability of blue destroying red before red reaches launch range can be computed. This computation is for a SAR guided missile. This computation and comparison is a good way to validate the analytical model. The condition that must be met for blue to destroy red before red reaches launch range is given by equation 25. **Z$_B$** or "blue range advantage" is given by equation 26. The probability that **Z$_B$** > 0 is the probability that blue destroys red before red reaches launch range.

$$\left(1 - \frac{\sin(\rho)}{\sin(\alpha_{MB})} \cdot \frac{V_C}{V_{CMB}}\right) \cdot (D_B - L_B) > D_R - L_R \quad (25)$$

$$Z_B = \left(1 - \frac{\sin(\rho)}{\sin(\alpha_{MB})} \cdot \frac{V_C}{V_{CMB}}\right) \cdot (D_B - L_B) - (D_R - L_R) > 0 \quad (26)$$

Because all random variables were assumed normally distributed the mean and variance of **Z$_B$** can be easily computed and are given by equation 27.

$$\mu_{ZB} = \left(1 - \frac{\sin(\rho)}{\sin(\alpha_{MB})} \cdot \frac{V_C}{V_{CMB}}\right) \cdot (d_B - l_B) - (d_R - l_R)$$

$$\sigma_{ZB}^2 = \left(1 - \frac{\sin(\rho)}{\sin(\alpha_{MB})} \cdot \frac{V_C}{V_{CMB}}\right)^2 \cdot (\sigma_{DB}^2 + \sigma_{LB}^2) + (\sigma_{DR}^2 + \sigma_{LR}^2) \quad (27)$$

For an example, a situation is constructed for which blue and red are equally matched. Generic values were chosen for input to the analytical model and can be seen on the left half of figure 5. The results can be seen on the right half of figure 5. Blue has a 12% probability of destroying red before red reaches launch range. This compares very well with reference 5, although not exactly because the model has been enhanced.

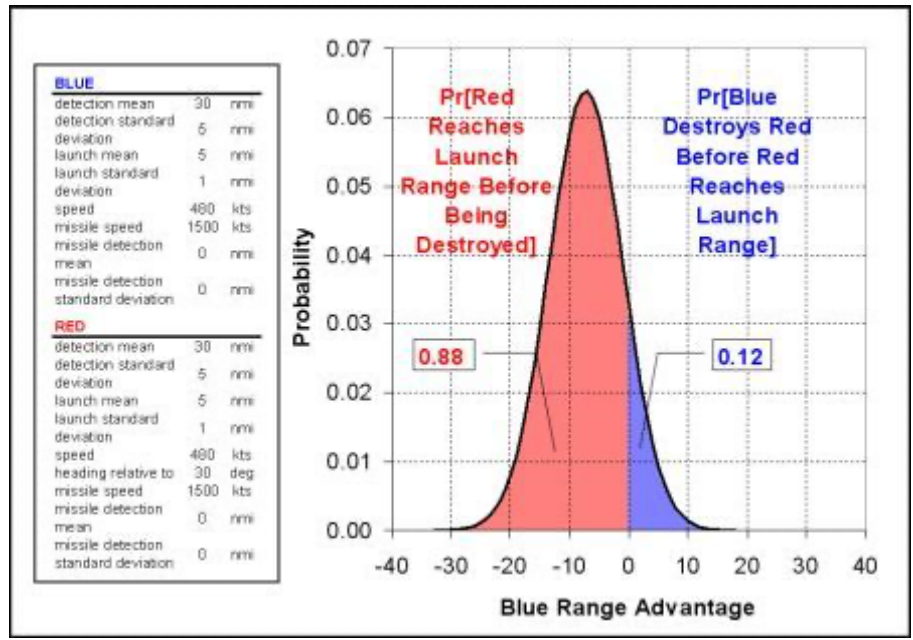

Figure 5.5

A similar analysis can be done with a slight enhancement to the above model for AR missiles. In this case there is an additional random variable, **D**$_{MB}$, to account for the range the missile seeker acquires the target. The extra random variable is added to equation 26 to make equation 28. The mean and standard deviation for equation 28 are given by equation 29.

$$Z_B = \left(1 - \frac{\sin(\rho)}{\sin(\alpha_{MB})} \cdot \frac{V_C}{V_{CMB}}\right) \cdot (D_B - L_B + D_{MB}) - (D_R - L_R) > 0$$
$$D_{MB} \sim N[d_{MB}, \sigma_{DMB}] \tag{28}$$

$$\mu_{ZB} = \left(1 - \frac{\sin(\rho)}{\sin(\alpha_{MB})} \cdot \frac{V_C}{V_{CMB}}\right) \cdot (d_B - l_B + d_{MB}) - (d_R - l_R)$$

$$\sigma_{ZB}^2 = \left(1 - \frac{\sin(\rho)}{\sin(\alpha_{MB})} \cdot \frac{V_C}{V_{CMB}}\right)^2 \cdot (\sigma_{DB}^2 + \sigma_{LB}^2 + \sigma_{DMB}^2) + (\sigma_{DR}^2 + \sigma_{LR}^2) \tag{29}$$

The same example as before is re-worked with the AR missile model. The results can be seen in figure 6. As would be expected, blue fairs better. Blue has a 28% probability of destroying red before red reaches launch range.

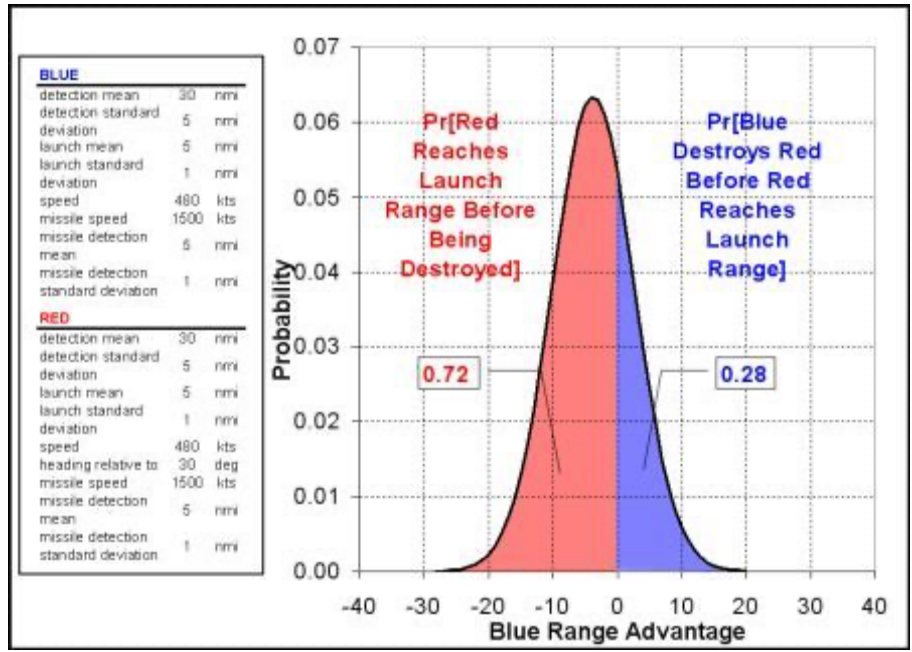

Figure 5.6

The above model can be further enhanced to represent a single missile dual. A blue "win" happens when the blue missile destroys red before the red missile seeker acquires. Conversely, a red "win" happens when the red missile destroys blue before the blue missile seeker acquires. A third outcome, a mutual kill, is possible. It happens when either side's missile acquires before the launcher is destroyed. The condition when blue wins is represented by equation 30 and the condition when red wins is represented by equation 31.

$$Z_B = \left(1 - \frac{\sin(\rho)}{\sin(\alpha_{MB})} \cdot \frac{V_C}{V_{CMB}}\right) \cdot (D_B - L_B) - \left(1 - \frac{\sin(\beta)}{\sin(\alpha_{MR})} \cdot \frac{V_C}{V_{CMR}}\right) \cdot (D_R - L_R + D_{MR}) < 0 \tag{30}$$

$$Z_R = \left(1 - \frac{\sin(\rho)}{\sin(\alpha_{MB})} \cdot \frac{V_C}{V_{CMB}}\right) \cdot (D_B - L_B + D_{MB}) - \left(1 - \frac{\sin(\beta)}{\sin(\alpha_{MR})} \cdot \frac{V_C}{V_{CMR}}\right) \cdot (D_R - L_R) < 0 \tag{31}$$

The mean and variance for equation 30 and equation 31 are computed in equation 32 and equation 33, respectively. The probability of a mutual kill is given by equation 34.

$$\mu_{ZB} = \left(1 - \frac{\sin(\rho)}{\sin(\alpha_{MB})} \cdot \frac{V_C}{V_{CMB}}\right) \cdot (d_B - l_B) - \left(1 - \frac{\sin(\beta)}{\sin(\alpha_{MR})} \cdot \frac{V_C}{V_{CMR}}\right) \cdot (d_R - l_R + d_{MR})$$

$$\sigma_{ZB}^2 = \left(1 - \frac{\sin(\rho)}{\sin(\alpha_{MB})} \cdot \frac{V_C}{V_{CMB}}\right)^2 \cdot (\sigma_{DB}^2 + \sigma_{LB}^2) + \left(1 - \frac{\sin(\beta)}{\sin(\alpha_{MR})} \cdot \frac{V_C}{V_{CMR}}\right)^2 \cdot (\sigma_{DR}^2 + \sigma_{LR}^2 + \sigma_{DMR}^2) \quad (32)$$

$$\mu_{ZR} = \left(1 - \frac{\sin(\rho)}{\sin(\alpha_{MB})} \cdot \frac{V_C}{V_{CMB}}\right) \cdot (d_B - l_B + d_{MB}) - \left(1 - \frac{\sin(\beta)}{\sin(\alpha_{MR})} \cdot \frac{V_C}{V_{CMR}}\right) \cdot (d_R - l_R)$$

$$\sigma_{ZR}^2 = \left(1 - \frac{\sin(\rho)}{\sin(\alpha_{MB})} \cdot \frac{V_C}{V_{CMB}}\right)^2 \cdot (\sigma_{DB}^2 + \sigma_{LB}^2 + \sigma_{DMB}^2) + \left(1 - \frac{\sin(\beta)}{\sin(\alpha_{MR})} \cdot \frac{V_C}{V_{CMR}}\right)^2 \cdot (\sigma_{DR}^2 + \sigma_{LR}^2) \quad (33)$$

$$P_{MK} = 1 - P(Z_B < 0) - P(Z_R < 0) \quad (34)$$

Again, a small example is presented to illustrate the analytical model. As before, blue and red have the same inputs and therefore are evenly matched. Figure 7 shows the results. As might be expected for an evenly matched pair, blue wins a quarter of the time, red wins a quarter of the time, and half the time the dual result in a mutual kill.

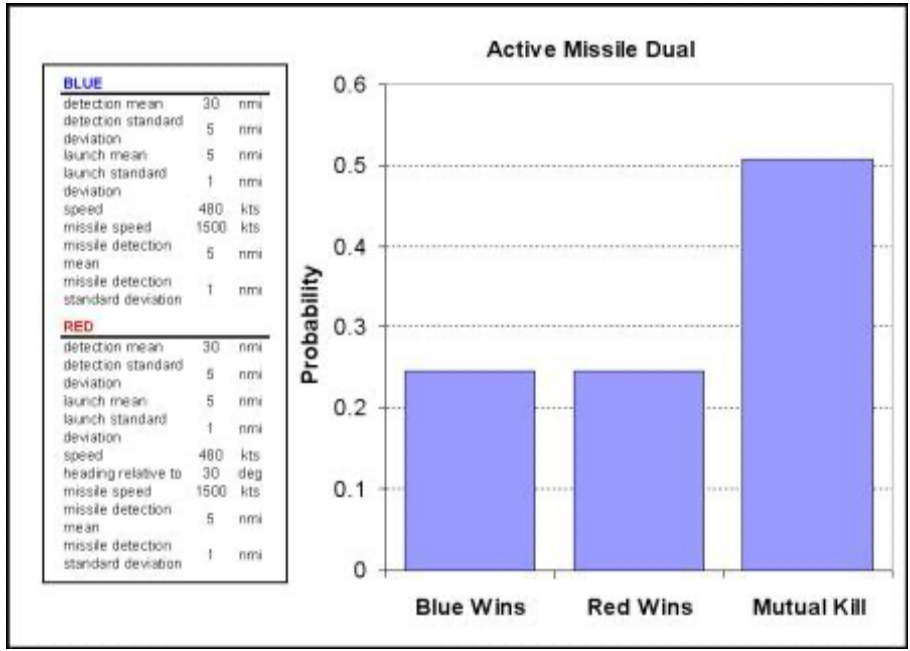

Figure 5.7

# APPENDIX A

# CODES

1. This code will check whether we can detect the missile or not. We are getting the object distance and its angle from the Arduino. The code is given below:

```cpp
#include<bits/stdc++.h>
using namespace std;

int main()
{
  // int xb[13]= {80,90,99,108,116,125,133,141,151,80,60,16,10};
   int xb[13]= {80,90,99,108,116,125,133,141,151,160,169,174,180};

  int yb[13]= {0,-2,-5,-9,-15,-18,-23,-29,-28,-25,-21,-20,-17};
   //int yb[13]= {0,-2,-5,-9,-15,-18,-23,-29,-28,-23,-25,-21,-28};

  float sin[13],cos[13],xf[13],yf[13];

  xf[0]=-70;
```

```c
    yf[0]=0;

    int i,vf=20;
    float ds=0;
    int flag =0;

    printf("| Time | xb | yb | xf | yf | Distance | Cos | Sin | \n");

    for (i=0; i<=12; i++)
    {

        ds=((xb[i]-xf[i])*(xb[i]-xf[i]))+((yb[i]-yf[i])*(yb[i]-yf[i]));
        ds=sqrt(ds);
        cos[i]=(xb[i]-xf[i])/ds;
        sin[i]=(yb[i]-yf[i])/ds;
            printf("| %d | %d | %d | %.2f | %.2f | %.2f | %.2f | %.2f | \n",i,xb[i],yb[i],xf[i],yf[i],ds,cos[i],sin[i]);

        if(ds<10){
           flag=1;
           break;
        }
        yf[i+1]=yf[i] + vf*sin[i];
        xf[i+1]=xf[i] + vf*cos[i];

    }

if(flag==1) printf("Target destroyed");
else printf("Target escaped");

    return 0;
}
```

2. Smoke Detection Process :

```cpp
#include<bits/stdc++.h>
using namespace std;

int redLed = 12;
int greenLed = 11;
int buzzer = 10;
int smokeA0 = A5;
// threshold value
int sensorThres = 400;

void setup() {
  pinMode(redLed, OUTPUT);
  pinMode(greenLed, OUTPUT);
  pinMode(buzzer, OUTPUT);
  pinMode(smokeA0, INPUT);
  Serial.begin(9600);
}

void loop() {
  int analogSensor = analogRead(smokeA0);

  Serial.print("Pin A0: ");
  Serial.println(analogSensor);
  // Checks if it has reached the threshold value
  if (analogSensor > sensorThres)
  {
    digitalWrite(redLed, HIGH);
    digitalWrite(greenLed, LOW);
    tone(buzzer, 1000, 200);
  }
  else
  {
    digitalWrite(redLed, LOW);
    digitalWrite(greenLed, HIGH);
    noTone(buzzer);
  }
  delay(100);
}
```

3. Aurdino Detection process :

```
// Includes the Servo library
#include <Servo.h>.
// Defines Tirg and Echo pins of the Ultrasonic Sensor
const int trigPin = 10;
const int echoPin = 11;
// Variables for the duration and the distance
long duration;
int distance;
Servo myServo; // Creates a servo object for controlling the servo motor
void setup() {
  pinMode(trigPin, OUTPUT); // Sets the trigPin as an Output
  pinMode(echoPin, INPUT); // Sets the echoPin as an Input
  Serial.begin(9600);
  myServo.attach(12); // Defines on which pin is the servo motor attached
}
void loop() {
  // rotates the servo motor from 15 to 165 degrees
  for(int i=15;i<=165;i++){
  myServo.write(i);
  delay(30);
  distance = calculateDistance();// Calls a function for calculating the distance measured by the Ultrasonic sensor for each degree

  Serial.print(i); // Sends the current degree into the Serial Port
  Serial.print(","); // Sends addition character right next to the previous value needed later in the Processing IDE for indexing
  Serial.print(distance); // Sends the distance value into the Serial Port
  Serial.print("."); // Sends addition character right next to the previous value needed later in the Processing IDE for indexing
  }
  // Repeats the previous lines from 165 to 15 degrees
  for(int i=165;i>15;i--){
  myServo.write(i);
```

```
  delay(30);
  distance = calculateDistance();
  Serial.print(i);
  Serial.print(",");
  Serial.print(distance);

 Serial.print(".");
  }
}
// Function for calculating the distance measured by the Ultrasonic sensor
int calculateDistance(){

  digitalWrite(trigPin, LOW);
  delayMicroseconds(2);
  // Sets the trigPin on HIGH state for 10 micro seconds
  digitalWrite(trigPin, HIGH);
  delayMicroseconds(10);
  digitalWrite(trigPin, LOW);
  duration = pulseIn(echoPin, HIGH); // Reads the echoPin, returns the sound wave travel time in microseconds
  distance= duration*0.034/2;
  return distance;
}
```

Note: This paper was written in 2018 and updated in 2024. It's a work in progress paper.